# EAGLE: Early Approximated-Gradient-based Learning rate Estimator

**Takumi Fujimoto†, Hiroaki Nishi‡**

†Faculty of Science and Technology, takmin@keio.jp

‡Center for Information and Computer Science, School of Science for Open and Environmental Systems, Keio University, west@keio.jp

## Abstract

We propose EAGLE update rule, a novel optimization method that accelerates loss convergence during the early stages of training by leveraging both current and previous step parameter and gradient values. The update algorithm estimates optimal parameters by computing the changes in parameters and gradients between consecutive training steps and leveraging the local curvature of the loss landscape derived from these changes. However, this update rule has potential instability, and to address that, we introduce an adaptive switching mechanism that dynamically selects between Adam and EAGLE update rules to enhance training stability. Experiments on standard benchmark datasets demonstrate that EAGLE optimizer, which combines this novel update rule with the switching mechanism achieves rapid training loss convergence with fewer epochs, compared to conventional optimization methods.

## 1 Introduction

### 1.1 Background

Deep learning has been widely adopted in various fields, including computer vision and natural language processing. In computer vision, the evolution of Convolutional Neural Network (CNN) [1] has achieved near-human recognition accuracy in tasks such as image classification and object detection. In natural language processing, while early approaches utilized Recurrent Neural Network (RNN) [2] and Long Short-Term Memory (LSTM) [3] networks—which addressed the long-term dependency problem—, Transformers [4] have become the dominant architecture in recent years. Transformers enable parallel computation and serve as the foundation for large language models such as Bidirectional Encoder Representations from Transformers (BERT) [5] and the GPT series [6], which demonstrate human-comparable capabilities in text generation and language understanding.

The performance of these deep learning models is quantitatively evaluated using task-specific metrics. superior models demonstrate high performance on these metrics, specifically showing minimal loss (error) between predicted and true values. Conversely, models with insufficient performance exhibit low prediction reliability, making practical applications challenging. Thus, obtaining high-performance models remains a crucial challenge in deep learning.

Essential to achieving high-performance models is the design of appropriate training processes. Model training is executed as an iterative process consisting of the following steps:

    (i)    Input data into the neural network to generate predictions

    (ii)   Calculate loss by inputting predicted and true values into a loss function (or error function)

    (iii)  Update model parameters to minimize the loss

    (iv)  Repeat the above steps until convergence criteria are met

Parameter optimization in deep learning models involves searching for parameter values that minimize loss through this process. The final model performance is determined by the parameter values obtained at the completion of training. While identifying optimal parameters would be straightforward if we could obtain the loss landscape for all parameters, modern deep learning models contain millions to billions of parameters, making complete loss landscape analysis computationally infeasible.

Optimizers in deep learning are algorithms designed to efficiently optimize parameters for such opaque loss functions. Efficient parameter updates by optimizers enhance model training efficiency, enabling faster model development and reduced computational resource requirements. The focus and objectives of optimizers have evolved over time. Early research primarily adopted simple optimization techniques such as Gradient Descent (GD) [7], which relied solely on gradient information. While theoretically straightforward, this approach faced challenges in convergence speed and computational efficiency, leading to the development of Stochastic Gradient Descent (SGD) [8] and the introduction of Momentum [9]. Subsequently, research focus shifted to adaptive learning rate adjustment. RMSprop [10] introduced mechanisms for adaptively adjusting learning rates for individual parameters, achieving efficient training. Notably, Adam [11] established a method combining the advantages of adaptive learning rate adjustment and momentum, demonstrating stable convergence generalization across various tasks. Recent trends, driven by the popularity of large-scale models, emphasize memory efficiency and computational cost reduction. Adafactor [12] reduces memory usage through low-rank approximation of second moments, while Lion [13] achieves computational cost reduction using update rules based solely on gradient sign information.

Thus, optimizer research has evolved from simple gradient information utilization to stable convergence through adaptive learning rate adjustment, and further to computational efficiency improvement, reflecting shifts in focus and objectives.

## 1.2 Objectives and Overview of the Proposed Method

As discussed in Section 1.1, research on optimizers in deep learning has progressed under various objectives, including convergence stability, computational efficiency, and memory efficiency. In this study, we propose EAGLE, a parameter optimization method focused on loss landscape characteristics, aiming to achieve both rapid loss convergence in early epochs and enhanced overall training stability. EAGLE features two main characteristics:

    (i)   **EAGLE Update Rule**:

        The update rule estimates optimal parameters by utilizing the local curvature of the loss function between consecutive steps, which is derived from the ratio of gradient changes to parameter changes in neural networks. This approach aims to accelerate loss convergence during the early stages of training.

    (ii)   **Adaptive Switching Mechanism**:

        EAGLE update rule has constraints related to extremely small gradient differences and the shape of the loss function. When these constraints are met, indicating that EAGLE update rule may not function appropriately, the optimizer switches to the established Adam update rule. This switching mechanism ensures overall training stability.

## 2   EAGLE UPDATE RULE

In this study, we describe EAGLE, a parameter update rule that utilizes the ratio of gradient changes to parameter changes in neural networks. Hereafter, we refer to the proposed update rule as "EAGLE update rule" and simply use "update" when discussing parameter updates. While the term "loss function" generally encompasses both "the relationship between loss and training steps" and "the relationship between loss and parameters," this paper uses it in the latter sense.

The mathematical formulation of EAGLE update rule is as follow:

$$\theta_{n+1} = \theta_n - \frac{\Delta\theta}{\Delta\frac{\partial L}{\partial \theta}} \cdot \frac{\partial L}{\partial \theta_n}$$

$$(\Delta\theta = \theta_n - \theta_{n-1}, \qquad \Delta\frac{\partial L}{\partial \theta} = \frac{\partial L}{\partial \theta_n} - \frac{\partial L}{\partial \theta_{n-1}})$$

where $\theta$ represents the neural network parameters, $\frac{\partial L}{\partial \theta}$ represents the gradient of the loss function $L$ with respect to $\theta$, $\Delta\theta \cdot \Delta(\partial L/\partial \theta)$ represents their respective changes, and $n$ represents the training step. EAGLE update rule utilizes information from both the current parameters $\theta_n$ and gradient $\partial L/\partial \theta_n$, as well as the previous step's parameters $\theta_{n-1}$ and gradient $\partial L/\partial \theta_{n-1}$. By calculating and taking the ratio of their differences during updates, we can estimate the curvature of the loss function between the current and previous steps. This curvature is a crucial characteristic indicating the trend of loss function changes, enabling updates in directions where gradient reduction is predicted. This curvature-aware approach allows for improved model training speed compared to conventional methods.

To verify the effectiveness of EAGLE update rule, let us consider a simple quadratic function as an example:

$$L = (\theta - 2)^2 + 2 = \theta^2 - 4\theta + 6$$

Taking the derivative with respect to θ:

$$\frac{\partial L}{\partial \theta} = \frac{\partial}{\partial \theta}(\theta^2 - 4\theta + 6) = 2\theta - 4$$

Hereafter, let $\partial L/\partial \theta_n = g_n$. We set the previous parameter $\theta_{n-1}$ and current parameter $\theta_n$ as follows, obtaining their respective gradients $g_{n-1}, g_n$:

$$\begin{cases} \theta_{n-1} = 10 \\ g_{n-1} = 2 \cdot 10 - 4 = 20 - 4 = 16 \end{cases}$$

$$\begin{cases} \theta_n = 8 \\ g_n = 2 \cdot 8 - 4 = 16 - 4 = 12 \end{cases}$$

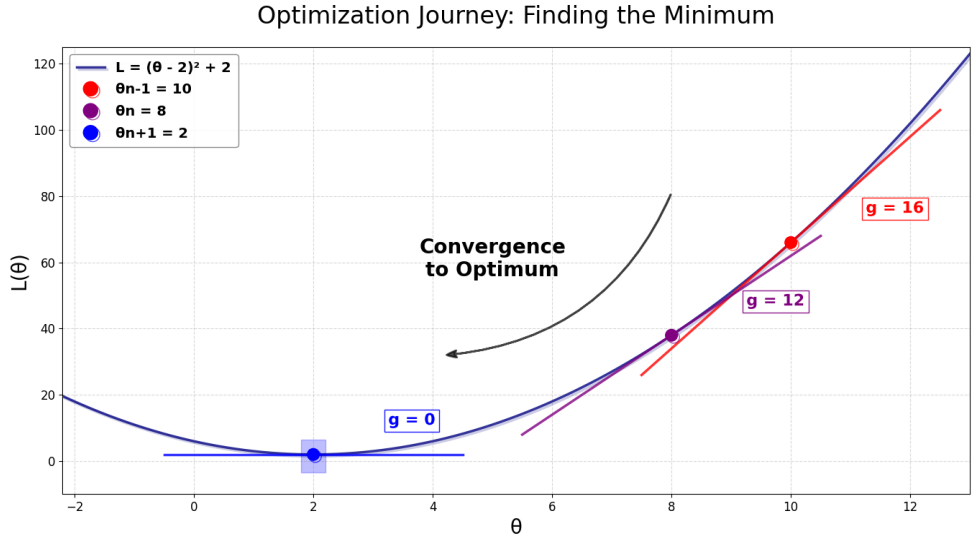

Figure 1: Visualization of the parameter update process in the example using the quadratic function $L = (\theta - 2)^2 + 2$. Three key points are shown: $\theta_{n-1} = 10$ (red) with gradient $g = 16$, $\theta_n = 8$ (purple) with gradient $g = 12$, and the estimated optimal parameter $\theta_{n+1} = 2$ (blue) with gradient $g = 0$.

Using these values, we estimate the optimal parameters with EAGLE update rule:

$$\theta_{n+1} = \theta_n - \frac{\theta_n - \theta_{n-1}}{g_n - g_{n-1}} g_n$$

$$= 8 - \frac{8 - 10}{12 - 16} \cdot 12$$

$$= 8 - \frac{(-2)}{(-4)} \cdot 12 = 8 - 6 = 2$$

Let us examine the parameter value $\theta_{n+1} = 2$ obtained through this update. At this point, the gradient $g_{n+1}$ in the original function $L = (\theta - 2)^2 + 2$ becomes zero ($g_{n+1} = 2 \cdot 2 - 4 = 0$), coinciding with the function's minimum point. This sequence is illustrated in Figure 1.

This result demonstrates that EAGLE update rule can estimate parameters that minimize loss in a single update when the loss function has a quadratic shape. Generalizing this finding suggests that EAGLE update rule may improve convergence speed in neural network training when the loss function is convex.

Furthermore, this update rule can be interpreted as a dynamic learning rate. While conventional optimization methods use fixed learning rates, in EAGLE update rule, the rate of change $\Delta\theta/\Delta g$ functions as an adaptive learning rate.

Stochastic Gradient Descent (SGD) uses the following update equation:

$$\theta_{n+1} = \theta_n - \alpha \cdot \frac{\partial L}{\partial \theta_n}$$

where $\alpha$ is a fixed learning rate. In contrast, EAGLE update rule uses the following update equation:

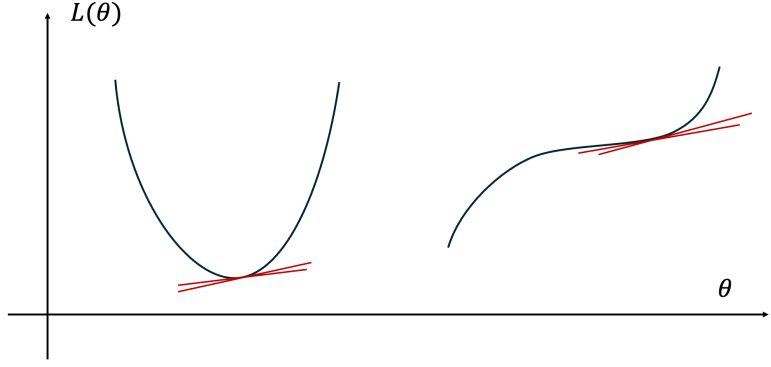

Figure 2: gradient changes causing update divergence

$$\theta_{n+1} = \theta_n - \beta \cdot \frac{\partial L}{\partial \theta_n}$$

$$\beta = \frac{\Delta \theta}{\Delta \frac{\partial L}{\partial \theta}} = \frac{\theta_n - \theta_{n-1}}{\frac{\partial L}{\partial \theta_n} - \frac{\partial L}{\partial \theta_{n-1}}}$$

Thus, EAGLE update rule can be viewed as adaptively adjusting the learning rate $\beta$ based on the respective changes in parameters and gradients.

## 3 Adaptive Switching Mechanism

EAGLE update rule proposed in Section 2 faces several practical challenges. To address these challenges, we introduce the adaptive switching mechanism for the update rule. Specifically, we adopt both Adam and EAGLE update rules, switching to Adam update rule in situations where EAGLE update rule is deemed ineffective. This switching mechanism is implemented to enhance training stability in practical neural networks. While we chose Adam among conventional optimization methods due to its widely recognized general performance with high training stability, essentially any optimizer could be adopted as an alternative. In the following sections, we describe two major practical challenges and the corresponding conditions for determining update rule switching.

### 3.1 Selection Based on Gradient Difference

The first challenge of EAGLE update rule is that when the gradient difference ratio $\Delta(\partial L/\partial \theta)$ in the denominator of the update term becomes extremely small, the update magnitude may diverge. This challenge can be expressed by the following relationship:

$$\Delta \frac{\partial L}{\partial \theta} \to 0$$

$$\therefore \quad \frac{\Delta \theta}{\Delta \frac{\partial L}{\partial \theta}} \to \pm \infty$$

$$\therefore \quad \theta_{n+1} = \theta_n - \frac{\Delta \theta}{\Delta \frac{\partial L}{\partial \theta}} \cdot \frac{\partial L}{\partial \theta_n} = \theta_n \mp \infty \to \mp \infty$$

This issue occurs primarily in two training processes, illustrated in Figure 2. First, while gradient differences show relatively large values in the early stages of training, as training progresses and parameters approach the minimum point, the gradient gradually decreases, causing gradient differences to converge to small values. Consequently, the gradient difference between consecutive steps becomes extremely small, potentially causing update magnitude divergence (left graph in Figure 2). Second, in locally flat regions of the loss function, the gradient maintains small, nearly constant values, similarly resulting in extremely small gradient differences that may cause update magnitude divergence (right graph in Figure2).

To address this challenge, we introduce an adaptive switching mechanism using the following conditional branching:

$$condition1: \left| \Delta \left( \frac{\partial L}{\partial \theta} \right) \right| < threshold$$

$$update\_rule = \begin{cases} Adam & (if\ condition1) \\ EAGLE & (otherwise) \end{cases}$$

where $threshold$ is a small positive threshold value. In this mechanism, Adam update rule is selected when the absolute value of the gradient difference is smaller than the $threshold$, and EAGLE update rule is selected otherwise. $threshold$ is a crucial hyperparameter. Setting an excessively large value biases the updates toward Adam, preventing the utilization of EAGLE update rule's advantages. Conversely, an excessively small value negates the intended function of avoiding division by extremely small values, making it impossible to suppress update magnitude divergence. Therefore, appropriate value setting is necessary. This switching enables flexible updates where Adam's superior stability is utilized in regions with minimal gradient changes, while EAGLE's rapid convergence is leveraged in regions with sufficient gradient changes.

## 3.2 Selection Based on Loss Landscape

The second challenge of EAGLE update rule is that optimal parameter estimation fails when the loss landscape exhibits locally convex regions. While EAGLE update rule is expected to function effectively in simple concave structures as shown in Section 2, it fails to estimate optimal parameters in locally convex regions.

To address this challenge, we classify loss landscape shapes based on the signs and variation patterns of second derivatives of gradients **(without directly computing them)**, analyzing EAGLE update rule's effectiveness in each case. As a result, we propose switching conditions for cases where the rule is considered ineffective.

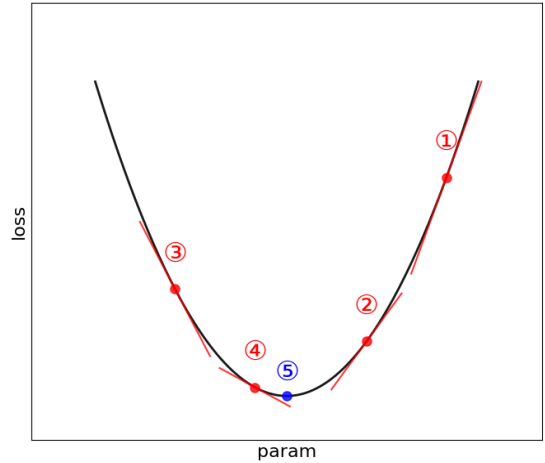

Figure 3: Representative Points and Their Gradients in the Loss Landscape (1)

- Case When Signs of Second Derivatives are Consistently Positive

    Figure 3 illustrates the typical shape of a loss landscape where second derivatives are consistently positive. In this case, the loss landscape exhibits global convexity, with gradients changing monotonically with respect to parameter changes. In Figure3, we define gradient characteristics at five representative points:

    **Point ①:** Point with large positive gradient

    **Point ②:** Point with small positive gradient

    **Point ③:** Point with large negative gradient

    **Point ④:** Point with small negative gradient

    **Point ⑤:** Point with zero gradient (global minimum)

    The transitions between these points can be classified into three basic patterns:

    **Transition 1-1:** Monotonic decrease in positive gradient (Point ① → Point ②)

    **Transition 1-2:** Monotonic increase in negative gradient (Point ③ → Point ④)

    **Transition 1-3:** Oscillation near optimal point (Point ② ↔ Point ④)

    To analyze convergence, we verify using the same simple quadratic function $L = (\theta - 2)^2 + 2$ as in Section 2. Since Transition 1-1 corresponds to the example discussed in Section 2 where effectiveness has already been confirmed, we analyze the convergence of Transition 1-2 and 1-3 below.

    ➤ Transition 1-2: Point ③ → Point ④

    We set the previous parameter $\theta_{n-1}$ and current parameter $\theta_n$ as follows, obtaining their respective gradients $g_{n-1}$, $g_n$:

$$\begin{cases} \theta_{n-1} = -8 \\ g_{n-1} = 2 \cdot (-8) - 4 = -16 - 4 = -20 \end{cases}$$

$$\begin{cases} \theta_n = -3 \\ g_n = 2 \cdot (-3) - 4 = -6 - 4 = -10 \end{cases}$$

Using these values, we estimate the optimal parameters:

$$\theta_{n+1} = \theta_n - \frac{\theta_n - \theta_{n-1}}{g_n - g_{n-1}} g_n$$

$$= -3 - \frac{-3 - (-8)}{-10 - (-20)} \cdot (-10)$$

$$= -3 - \frac{5}{10} \cdot (-10) = -3 + 5 = 2$$

As shown in Section 2, the obtained update value $\theta_{n+1} = 2$ is the parameter value that minimizes the function, indicating successful optimal parameter estimation. Therefore, the effectiveness of EAGLE update rule is confirmed in this specific example, suggesting its effectiveness in regions where negative gradients monotonically increase in actual neural networks.

➢   Transition 1-3: Point ② ↔ Point ④

Similarly, we set the previous parameter $\theta_{n-1}$ and current parameter $\theta_n$ as follows, obtaining their respective gradients $g_{n-1}$, $g_n$:

$$\begin{cases} \theta_{n-1} = -1 \\ g_{n-1} = 2 \cdot (-1) - 4 = -2 - 4 = -6 \end{cases}$$

$$\begin{cases} \theta_n = 5 \\ g_n = 2 \cdot 5 - 4 = 10 - 4 = 6 \end{cases}$$

Using these values, we estimate the optimal parameters:

$$\theta_{n+1} = \theta_n - \frac{\theta_n - \theta_{n-1}}{g_n - g_{n-1}} g_n$$

$$= 5 - \frac{5 - (-1)}{6 - (-6)} \cdot 6$$

$$= 5 - \frac{6}{12} \cdot 6 = 5 - 3 = 2$$

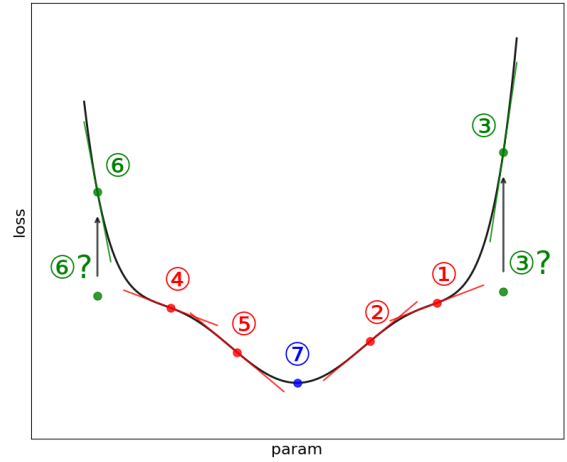

Figure 4: Representative Points and Their Gradients in the Loss Landscape (2)

Similarly, $\theta_{n+1} = 2$ is the parameter value that minimizes the function, indicating successful optimal parameter estimation. Therefore, this demonstrates that EAGLE update rule is effective even in regions where gradient signs change near the optimal point.

From the above analysis, we have shown that when the loss landscape has global convexity with monotonic gradient changes, EAGLE update rule tends to converge to the global minimum across all basic transition patterns. This confirms the effectiveness of EAGLE update rule.

- Case When Signs of Second Derivatives Change

Figure 4 illustrates a more complex case where signs of second derivatives change. In this case, the loss landscape has locally convex and concave structures, with gradients changing in complex ways with respect to parameter changes. In Figure 4, we define seven characteristic points:

**Point ①:**    Point with small positive gradient

**Point ②:**    Point with large positive gradient

**Point ③:**    Incorrect estimation point

**Point ④:**    Point with small negative gradient

**Point ⑤:**    Point with large negative gradient

**Point ⑥:**    Incorrect estimation point

**Point ⑦:**    Point with zero gradient (global minimum)

Under this situation, there exist two new transition patterns that were not observed in the "case when signs of second derivatives are consistently positive":

**Transition 2-1:**    Increase in positive gradient (Point ① → Point ②)

**Transition 2-2:**    Decrease in negative gradient (Point ④ → Point ⑤)

Table 1: EAGLE Effectiveness for Each Transition Pattern

| Transition Pattern | $\nabla L(n-1)$ | $\nabla L(n)$ | $\Delta \nabla L$ | EAGLE Effectiveness |
|---|---|---|---|---|
| Transition 1-1 | Large positive | Small positive | Negative | Effective |
| Transition 1-2 | Large negative | Small negative | Positive | Effective |
| Transition 1-3 | Positive (Negative) | Negative (Positive) | Negative (Positive) | Effective |
| Transition 2-1 | Small positive | Large positive | Positive | Ineffective |
| Transition 2-2 | Small negative | Large negative | Negative | Ineffective |

In these transitions, we observe a phenomenon where the absolute value of gradients increases between consecutive steps. This phenomenon suggests that the loss landscape may be locally convex in the region between updated parameters. While EAGLE update rule is designed to converge gradients to zero and works effectively in concave regions where zero gradients correspond to global minima, it performs inappropriate updates in convex regions where zero gradients do not correspond to minima, resulting in increased loss without gradient reduction. Specifically, in Figure 4, Points ③? and ⑥? are estimated as parameters where gradients converge to zero in Transitions 2-1 and 2-2 respectively, but their actual losses and gradients at Points ③ and ⑥ indicate inappropriate updates. Therefore, Adam update rule should be selected when the absolute value of gradients increases after updates.

Conversely, EAGLE update rule remains effective for transitions in regions where gradient signs change (Point ② ↔ Point ⑤). These transitions occur near the global minimum where the shape is concave, allowing EAGLE update rule to function effectively. This convergence characteristic is similar to what was observed in Transition 1-3 of the "case when second derivatives are consistently positive."

These analysis results, focusing on the relationship between gradients and their changes in sign between consecutive steps, are summarized in Table 1, where $\nabla L$ represents the gradient of the loss landscape with respect to parameters, n indicates the step number, and $\Delta \nabla L$ represents the change in gradients between consecutive steps. From this Table 1, EAGLE update rule is confirmed to be ineffective when the following two conditions are simultaneously satisfied:

**Condition 1:** Consistency of gradient signs between consecutive steps: $\nabla L(n-1) * \nabla L(n) \geq 0$

**Condition 2:** Consistency of gradient signs between current gradient and gradient difference: $\nabla L(n) * \Delta \nabla L \geq 0$

To address inappropriate updates under these conditions, we introduce an adaptive switching mechanism using the following conditional branching:

$$condition2: (\nabla L(n-1) * \nabla L(n) \geq 0) \wedge (L(n) * \Delta \nabla L \geq 0)$$

$$update\_rule = \begin{cases} Adam & (if\ condition2) \\ EAGLE & (otherwise) \end{cases}$$

In this mechanism, Adam update rule is selected when the loss landscape shape is deemed unsuitable for EAGLE update rule, and EAGLE update rule is selected otherwise. Through this adaptive switching, even in cases where the loss landscape is not necessarily concave, we achieve both training stability and convergence by combining Adam update rule's superior stability with EAGLE update rule's rapid convergence.

Furthermore, the above switching mechanism simultaneously resolves an implementation challenge that EAGLE faces in the first update step. While EAGLE update rule requires gradient history indicating the gradient information from the previous step, this gradient history does not exist in the first step immediately after training begins. Therefore, the first update must use Adam update rule rather than EAGLE update rule, necessitating a separate branching process to use Adam update rule. However, by defining the previous gradient as zero in the first update, we can make the above switching mechanism function as a solution.

Let us calculate the gradient change by defining the gradient at the first update as a real number $k$ and the previous gradient as zero:

$$\nabla L(1) = k, \qquad \nabla L(0) = 0, \qquad \Delta \nabla L = \nabla L(1) - \nabla L(0) = k$$

Substituting these values into $condition2$:

$$\therefore \begin{cases} \nabla L(n-1) * \nabla L(n) = \nabla L(0) * \nabla L(1) = 0 * k = 0 \geq 0 \\ L(n) * \Delta \nabla L = L(1) * \Delta \nabla L = k * k = k^2 \geq 0 \end{cases}$$

Consequently, since $condition2$ holds true for any real number $k$, the update rule necessarily becomes Adam.

$$\therefore \ update\_rule = Adam \quad (\because \ \forall k \in R: condition2 = true)$$

This eliminates the need for separate branching logic for the first update step where gradient history does not exist. Additionally, from the second step onward, the necessary gradient history has been accumulated, enabling appropriate updates using EAGLE update rule. Thus, the switching mechanism based on gradient sign relationships functions not only as a mere update rule switching mechanism but also as a solution to the history dependency challenge in EAGLE update rule's first update.

### 3.3 Integration of Switching Mechanisms

Based on the above analysis, we integrate the switching mechanisms for update rules. $condition1$ based on gradient difference threshold shown in Section 3.1 and $condition2$ based on loss landscape shape shown in Section 3.2 are both mechanisms for determining switches to Adam update rule. Therefore, we combine these conditions as a logical OR and formulate them as a single selection equation:

$$condition1: \ \left| \Delta \left( \frac{\partial L}{\partial \theta} \right) \right| < threshold$$

$$condition2: (\nabla L(n-1) * \nabla L(n) \ \geq \ 0) \ \wedge \ (L(n) * \Delta \nabla L \ \geq \ 0)$$

$$condition: condition1 \ \vee \ condition2$$

$$update\_rule = \begin{cases} Adam & (if \ condition) \\ EAGLE & (otherwise) \end{cases}$$

This integrated switching mechanism enables consistent handling of both challenges—update magnitude divergence due to gradient difference minimization and inappropriate updates due to locally convex loss landscape shapes—within a unified framework.

## 4 PSEUDO-CODE IMPLEMENTATION

The proposal organized as pseudo-code is shown below.

---

**Algorithm**: EAGLE's update rule

---

1:   **Inputs**:
2:     $\alpha \in \mathbb{R}$: Learning rate
3:     $\beta_1, \beta_2 \in [0, 1)$: Exponential moving average decay rates
4:     $\varepsilon \in \mathbb{R}$: Small constant for numerical stability
5:     $threshold \in \mathbb{R}$: Gradient change threshold
6:     $\theta_0 \in \mathbb{R}^n$: Initial value of parameter vector
7:     $f(\theta)$: Objective function (loss function)
8:   **Initialize**:
9:     // Initialize states
10:     $n \leftarrow 0$                   // Store 0 in step count
11:     $m_0 \leftarrow 0$                // Store 0 in gradient mean
12:     $v_0 \leftarrow 0$                 // Store 0 in gradient variance
13:     $param_{prev} \leftarrow \theta_0$       // Store initial value $\theta_0$ in previous parameter
14:     $grad_{prev} \leftarrow 0$          // Store 0 in previous gradient
15:   **while** $\theta_{n+1}$ not converged **do**:
16:     // Variable definition
17:     $g_n \leftarrow \nabla L(\theta_n)$
18:     $param_{curr} \leftarrow \theta_n$
19:     $grad_{curr} \leftarrow g_n$
20:
21:     // Calculate changes in parameters and gradients
22:     $\Delta param \leftarrow param_{curr} - param_{prev}$
23:     $\Delta grad \leftarrow grad_{curr} - grad_{prev}$
24:
25:     // Calculate Adam momentum update and bias correction
26:     $m_n \leftarrow \beta_1 m_{n-1} + (1 - \beta_1)grad_{curr}, \ v_n \leftarrow \beta_2 v_{n-1} + (1 - \beta_2)grad_{curr}^2$
27:     $\hat{m}_n \leftarrow m_n/(1 - \beta_1^n), \ \hat{v}_n \leftarrow v_n/(1 - \beta_1^n)$
28:
29:     // Define update equations
30:     $adam_{update} \leftarrow \alpha \cdot \hat{m}_n/(\sqrt{\hat{v}_n} + \varepsilon)$
31:     $adaeagle_{update} \leftarrow grad_{curr} \cdot (\Delta param/\Delta grad)$
32:
33:     // Define branching conditions

34:     $condition_{lossfunc\_shape} \leftarrow (grad_{prev} \cdot grad_{curr} \geq 0) \land (grad_{curr} \cdot \Delta grad \geq 0)$

35:     $condition_{grad\_diff} \leftarrow (|\Delta grad| < threshold)$

36:     $condition_{adam} \leftarrow condition_{lossfunc\_shape} \lor condition_{grad\_diff}$

37:

38:     // Execute update

39:     **if** $condition_{adam}$ **then**

40:         $\theta_{n+1} \leftarrow \theta_n - adam_{update}$

41:     **else**

42:         $\theta_{n+1} \leftarrow \theta_n - adaeagle_{update}$

43:

44:     // Update each state

45:     $param_{prev} \leftarrow param_{curr}$

46:     $grad_{prev} \leftarrow grad_{curr}$

47:     $n+1 \leftarrow (n+1)+1$

48: **end while**

49: **return** $\theta_{n+1}$         // Update parameters to θ_(n+1)

## 5 RELATED WORK: ADAM

Adam (Adaptive Moment Estimation) [11], proposed by Kingma et al. in 2015, is an optimization method that combines the advantages of AdaDelta [14], Momentum [9], and RMSprop [10]. It is expressed by the following update equations:

$$m_n = \beta_1 m_{n-1} + (1 - \beta_1)\nabla L(\theta_n) \qquad (m_0 = 0)$$

$$v_n = \beta_2 v_{n-1} + (1 - \beta_2)\{\nabla L(\theta_n)\}^2 \qquad (v_0 = 0)$$

$$\hat{m}_n = \frac{m_n}{1 - \beta_1^n} \ , \qquad \hat{v}_n = \frac{v_n}{1 - \beta_2^n}$$

$$\theta_{n+1} = \theta_n - \eta \frac{\hat{m}_n}{\sqrt{\hat{v}_n} + \epsilon}$$

- First-Order Moment Calculation

  The first-order moment (mean) incorporates the Momentum concept similar to AdaDelta, calculating the moving average of gradients to consider the directional trends of past gradients and appropriately determine the update direction. This mechanism enables stable updates even with significant gradient changes.

- Second-Order Moment Calculation

  The second-order moment (variance) also incorporates the RMSprop concept similar to AdaDelta, calculating the moving average of squared gradient values to adaptively adjust learning rates individually for each parameter. This mechanism enables efficient learning based on parameter importance and update necessity.

- Bias Correction

  While Adam has relatively many hyperparameters, they exhibit high robustness, and their default values ($\beta_1 = 0.9$, $\beta_2 = 0.999$, $\varepsilon = 1e - 8$) function well across a wide range of tasks. Consequently, it is widely used as the de facto standard in deep learning.

## 6 EXPERIMENTS

In this chapter, we evaluate the performance of our proposed method on multi-layer neural networks and convolutional neural networks, analyze the local shape of loss functions in the datasets and network models used in Sections 6.1 and 6.2, and examine the usage ratio of EAGLE update rule with the switching mechanism introduced for practical extensions. For all implementation evaluations, we adopted a single machine equipped with an NVIDIA A100 80GB PCIe.

### 6.1 EXPERIMENT: MULTI-LAYER NEURAL NETWORKS

#### 6.1.1 EXPERIMENTAL SETTINGS

In this experiment, we evaluate the performance of EAGLE based on our research objectives: rapid loss convergence in the early epochs of training and improved overall training stability. We adopted datasets from the UCI Machine Learning Repository: the Iris and Wine datasets. The Iris dataset is used for classifying three types of iris species based on four features (sepal length, sepal width, petal length, petal width). We adopt a three-layer neural network (input layer: 4, hidden layer: 25, output layer: 3) as our model. The Wine dataset is used for classifying wine types based on their chemical composition. It predicts which of three types a wine belongs to using 13 features (alcohol, malic acid, ash, alkalinity of ash, magnesium, total phenols, flavonoids, nonflavanoid phenols, proanthocyanins, color intensity, hue, OD280/OD315 of diluted wines, proline). We adopt a three-layer neural network (input layer: 13, hidden layer: 15, output layer: 3) as our model.

The gradient difference threshold in the switching mechanism was set to 0.0005 through preliminary experiments. As comparison baselines, we selected conventional optimizers: SGD with momentum and Adam, and advanced optimizers: RAdam, Lion, and Sophia. In the following sections, we refer to SGD with momentum simply as SGD.

To ensure reproducibility and reliability, we conducted independent experiments using 10 randomly selected seed values from 1 to 10,000. We evaluated the experimental results using the following three evaluation metrics:

**Evaluation Metrics 1**: Time-series visualization of mean and standard deviation for Training Loss, Training Accuracy, Test Loss, and Test Accuracy across all seeds (shown in top and middle panels). The standard deviation is represented by shaded regions around each curve.

**Evaluation Metrics 2**: Early-stage learning progression graphs showing Training Loss and Training Accuracy, extracted from the above visualizations (shown in bottom panels).

**Evaluation Metrics 3**: Comparative analysis of Training Loss changes across optimizers

at two-epoch intervals during the early learning phase.

### 6.1.2 RESULTS AND DISCUSSION

First, we present the experimental results using the Iris dataset in Figure 5,6 and Table 2,3.

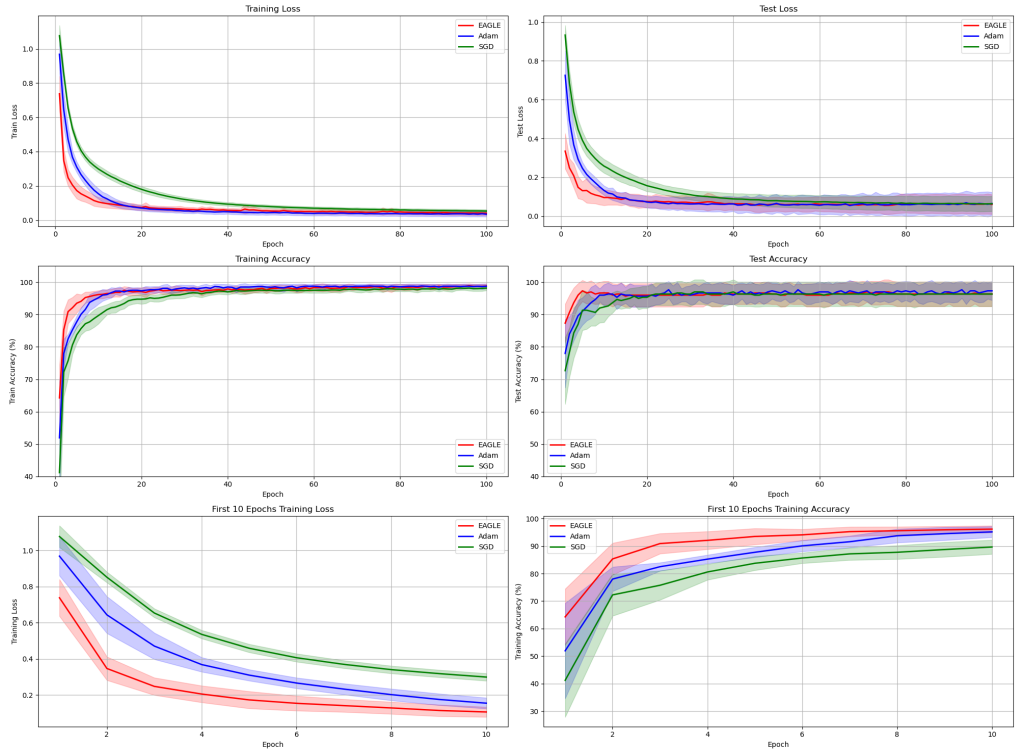

Figure 5: Comparison with Conventional Optimizers: Evaluation Metrics 1 and 2 (Iris)

Table 2: Comparison with Conventional Optimizers: Evaluation Metrics 3 (Iris)

|        | 2      | 4      | 6      | 8      | 10     |
|--------|--------|--------|--------|--------|--------|
| EAGLE  | **0.3463** | **0.2049** | **0.1536** | **0.1280** | **0.1058** |
| Adam   | 0.6431 | 0.3678 | 0.2657 | 0.2016 | 0.1543 |
| SGD    | 0.8513 | 0.5350 | 0.4058 | 0.3399 | 0.2988 |

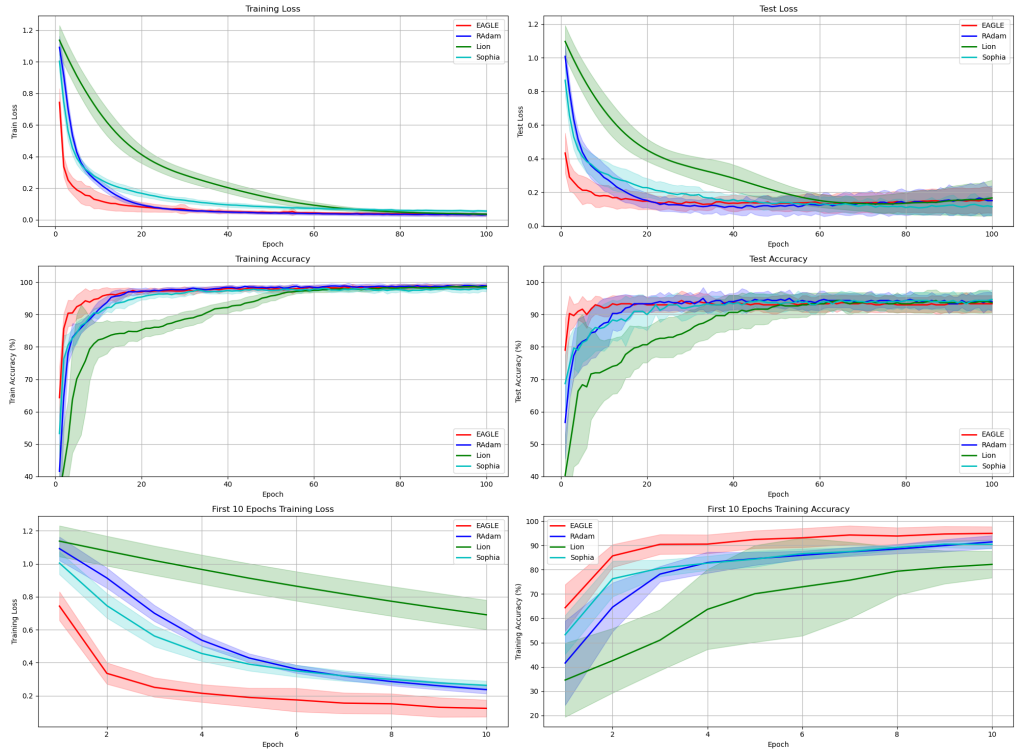

Figure 6: Comparison with Advanced Optimizers: Evaluation Metrics 1 and 2 (Iris)

Table 3: Comparison with Advanced Optimizers: Evaluation Metrics 3 (Iris)

|        | 2      | 4      | 6      | 8      | 10     |
|--------|--------|--------|--------|--------|--------|
| EAGLE  | **0.3357** | **0.2150** | **0.1747** | **0.1515** | **0.1236** |
| RAdam  | 0.9139 | 0.5368 | 0.3614 | 0.2860 | 0.2372 |
| Lion   | 1.0778 | 0.9658 | 0.8640 | 0.7731 | 0.6910 |
| Sophia | 0.7463 | 0.4562 | 0.3504 | 0.2993 | 0.2631 |

Next, we present the experimental results using the Wine dataset in Figure 7,8 and Table 4,5.

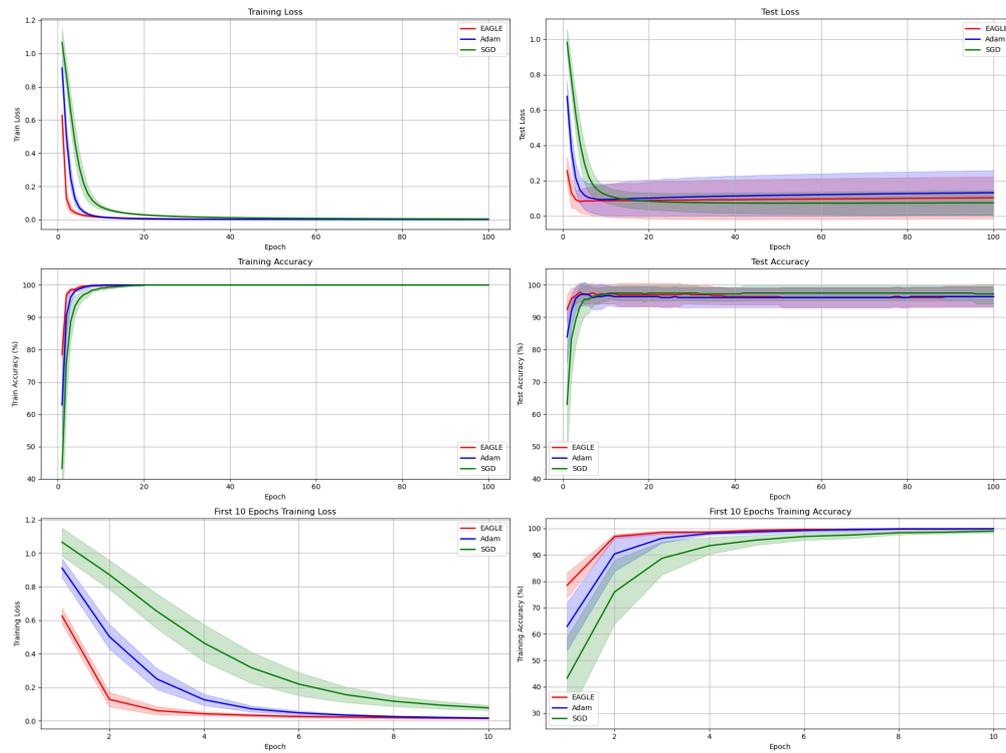

Figure 7: Comparison with Conventional Optimizers: Evaluation Metrics 1 and 2 (Wine)

Table 4: Comparison with Conventional Optimizers: Evaluation Metrics 3 (Wine)

|  | 2 | 4 | 6 | 8 | 10 |
|---|---|---|---|---|---|
| EAGLE | **0.1277** | **0.0417** | **0.0253** | **0.0187** | **0.0149** |
| Adam | 0.5023 | 0.1256 | 0.0475 | 0.0249 | 0.0161 |
| SGD | 0.8723 | 0.4638 | 0.2191 | 0.1169 | 0.0771 |

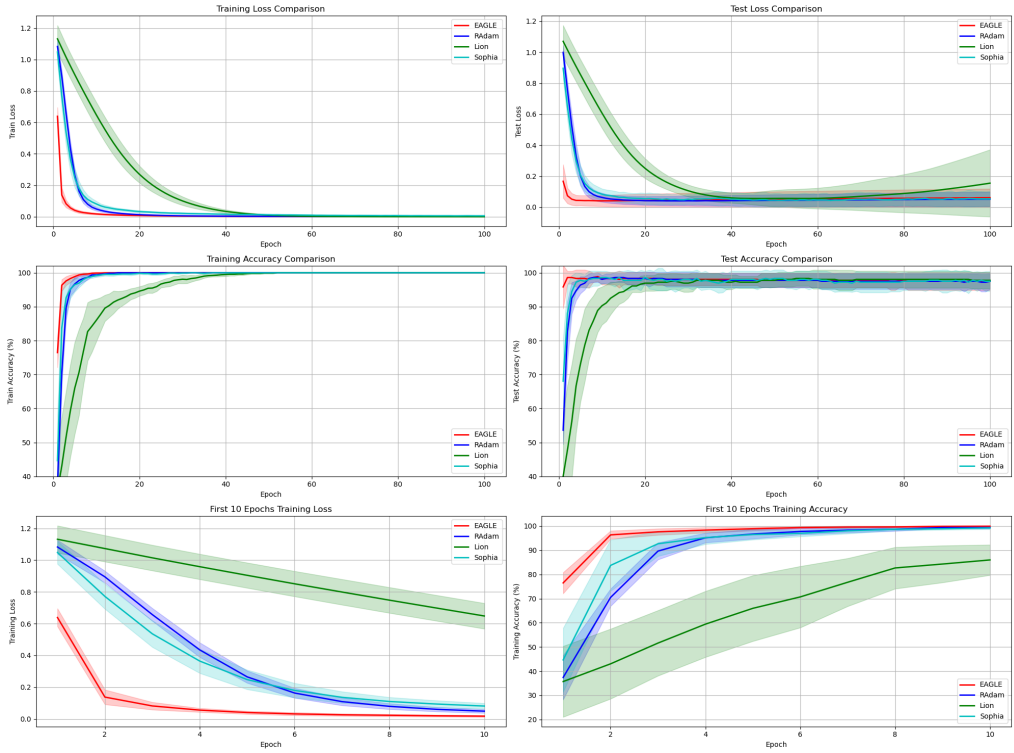

Figure 8: Comparison with Advanced Optimizers: Evaluation Metrics 1 and 2 (Wine)

Table 5: Comparison with Advanced Optimizers: Evaluation Metrics 3 (Wine)

|  | 2 | 4 | 6 | 8 | 10 |
|---|---|---|---|---|---|
| EAGLE | **0.1378** | **0.0557** | **0.0318** | **0.0231** | **0.0177** |
| RAdam | 0.8950 | 0.4349 | 0.1633 | 0.0794 | 0.0491 |
| Lion | 1.0731 | 0.9589 | 0.8510 | 0.7481 | 0.6485 |
| Sophia | 0.7719 | 0.3645 | 0.1798 | 0.1100 | 0.0813 |

We summarize the experimental results. For the Iris dataset, as shown in Figures 5, 6 and Tables 2, 3, the training loss values are consistently lower than all other optimizers throughout epochs 2 to 10. Notably, the training loss at epoch 2 shows approximately 46% lower value compared to Adam, which achieved the second-best performance. For the Wine dataset, as shown in Figures 7, 8 and Tables 4, 5, similar to the Iris dataset, the training loss values are consistently lower than all other optimizers throughout epochs 2 to 10. However, the superiority is even more pronounced compared to the Iris dataset, with the training loss at epoch 2 showing approximately 75% lower value compared to Adam, which achieved the second-best performance.

Based on these results, we can conclude that our proposed EAGLE optimizer has achieved our research objectives: rapid loss convergence in the early epochs while maintaining overall training stability.

However, training loss at the completion of learning is also an important metric for comprehensive evaluation of deep learning model performance. Therefore, we next compare the training loss values at epoch 100, where learning is completed, across all optimizers in Tables 6 and 7.

Table 6: Comparison of Training Loss at Epoch 100 with Conventional Optimizers

|       | Iris   | Wine   |
|-------|--------|--------|
| EAGLE | 0.0406 | 0.0003 |
| Adam  | **0.0345** | **0.0002** |
| SGD   | 0.0531 | 0.0039 |

Table 7: Comparison of Training Loss at Epoch 100 with Advanced Optimizers

|        | Iris   | Wine   |
|--------|--------|--------|
| EAGLE  | 0.0336 | 0.0004 |
| RAdam  | **0.0327** | 0.0005 |
| Lion   | 0.0354 | **0.0000** |
| Sophia | 0.0541 | 0.0059 |

As shown in Tables 6 and 7, EAGLE does not demonstrate the best performance in terms of training loss at the completion of learning (epoch 100). Specifically, for the Iris dataset, EAGLE's training loss values (0.0406, 0.0336) are higher compared to Adam and RAdam (0.0345, 0.0327). Similarly, for the Wine dataset, EAGLE's training loss values (0.0003, 0.0004) are higher compared to Adam and Lion (0.0002, 0.0000). Therefore, while EAGLE achieves rapid loss convergence in the early stages of training, the initial loss value differences diminish as training progresses, ultimately resulting in higher loss values for EAGLE. Since the local shape of loss functions may be a contributing factor to this inferior final convergence performance, we analyze this aspect in Section 6.3.

### 6.2 EXPERIMENT: CONVOLUTIONAL NEURAL NETWORKS

#### 6.2.1 EXPERIMENTAL SETTINGS

We conduct the same experiments as in Section 6.1 using the MNIST dataset to evaluate the performance of EAGLE. The MNIST dataset is used for recognizing handwritten digits from 0 to 9. Each digit is represented as a 28×28 pixel grayscale image. Each pixel has a brightness value from 0 to 255, which is normalized to the range [0,1] by dividing by 255 as preprocessing. We adopt a convolutional neural network (CNN) that combines two convolutional blocks and two fully connected layers. Each convolutional block includes a convolutional layer (kernel size 3×3), batch normalization layer, ReLU activation function, and max pooling layer (kernel size 2×2). The first fully connected layer has 3,136 (64×7×7) inputs and 128 outputs, followed by a batch normalization layer and ReLU activation function, while the second layer performs 10-class classification.

To ensure reproducibility and reliability, we conducted independent experiments using 5 randomly selected seed values from 1 to 10,000. For evaluation of the experimental results, we used Evaluation Metrics 1 and 3 from Section 6.1. However, in Metric 3, we present the training loss changes at one-epoch intervals during the early stages of learning.

#### 6.2.2 RESULTS AND DISCUSSION

we present the experimental results using the MNIST dataset in Figure 9,10 and Table 8,9.

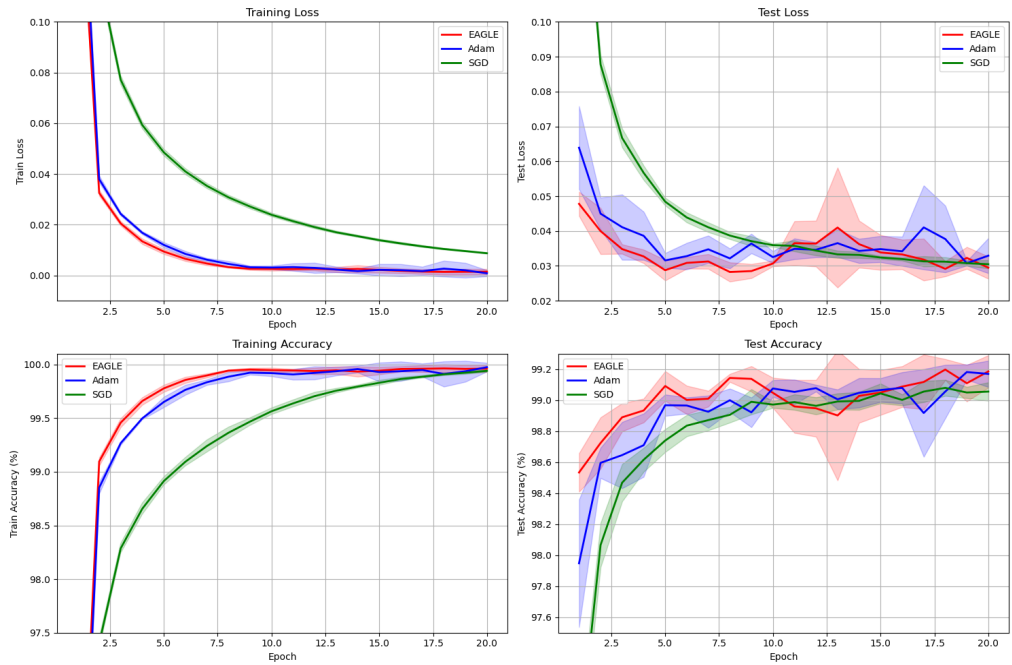

Figure 9: Comparison with Conventional Optimizers: Evaluation Metrics 1

Table 8: Comparison with Conventional Optimizers: Evaluation Metrics 3

|  | 1 | 2 | 3 | 4 |
|---|---|---|---|---|
| EAGLE | **0.1717** | **0.0325** | **0.0205** | **0.0134** |
| Adam | 0.1939 | 0.0380 | 0.0242 | 0.0168 |
| SGD | 0.5121 | 0.1177 | 0.0771 | 0.0593 |

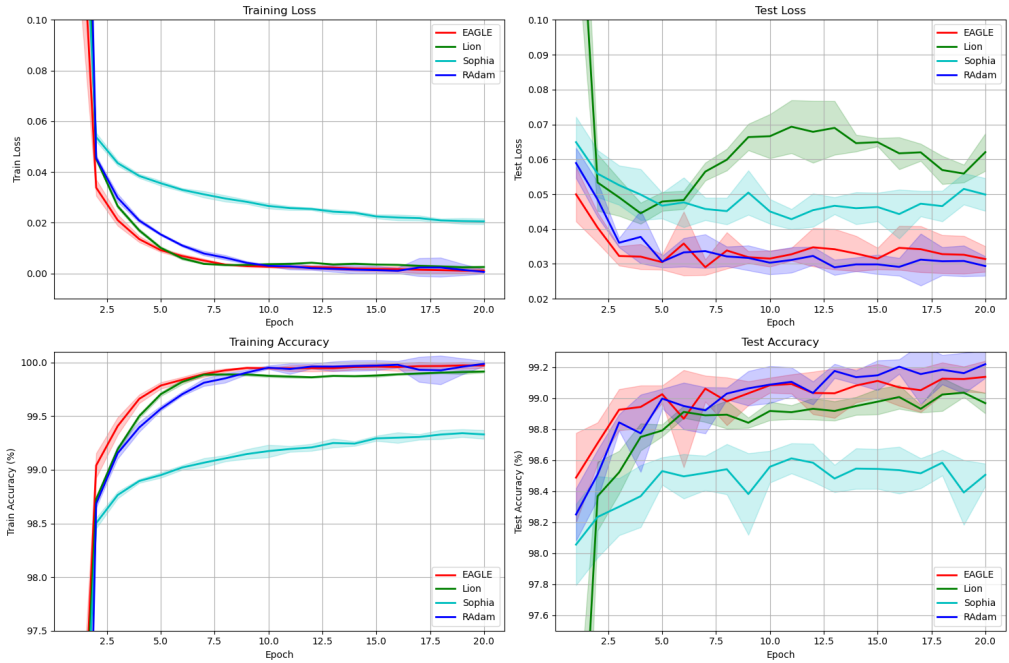

Figure 10: Comparison with Advanced Optimizers: Evaluation Metrics 1

Table 9: Comparison with Advanced Optimizers: Evaluation Metrics 3

|  | 1 | 2 | 3 | 4 |
|---|---|---|---|---|
| EAGLE | **0.1956** | **0.0338** | **0.0210** | **0.0135** |
| RAdam | 0.3922 | 0.0455 | 0.0298 | 0.0208 |
| Lion | 0.2883 | 0.0452 | 0.0265 | 0.0170 |
| Sophia | 0.2200 | 0.0537 | 0.0435 | 0.0384 |

As shown in Figures 9, 10 and Tables 8, 9, similar to the Iris and Wine datasets used in Section 6.1, EAGLE achieves the fastest training loss convergence compared to all other optimizers throughout epochs 1 to 4 for the MNIST dataset and its CNN model. However, the significant early-stage loss reduction observed in the Iris and Wine datasets was not apparent in these results. Since the local shape of loss functions may be a contributing factor to this result, we analyze this aspect in Section 6.3.

## 6.3 Experiment: Analysis of Loss Landscape

### 6.3.1 Experimental Settings

This research assumes that EAGLE update rule functions effectively when the loss function shape can be approximated as convex. Therefore, to verify this convexity property of loss functions, we experimentally analyzed the shape of loss functions using the Iris, Wine, and MNIST datasets and their corresponding models from Sections 6.1 and 6.2. The analysis procedure is described below:

1. **Model Training and Reference Point Setting**

   i. Train the neural network model

   ii. Set the optimal parameter values obtained by the optimizer (EAGLE in this experiment) as reference points

2. **Parameter Sampling**

   i. Sample a subset of parameters considering computational cost, rather than all parameters

   ii. Iris: Randomly select 50 parameters from input layer→hidden layer, 40 parameters from hidden layer→output layer

   iii. Wine: Randomly select 60 parameters from input layer→hidden layer, 30 parameters from hidden layer→output layer

   iv. MNIST: For convolutional layers, randomly select 20 parameters for each of the two blocks; for fully connected layers, randomly select 20 parameters for each of the two layers

3. **Sensitivity Analysis for Each Parameter**

   i. Fix all parameter values at their reference points

   ii. Select one parameter from the sampled set and vary it within ±5 range of the reference point, calculating loss values (with 1000-point uniform sampling, resulting in parameter intervals of 0.01)

   iii. Return the varied parameter to its reference point

   iv. Repeat steps ii and iii for all parameters selected in step 2

4. **Results Visualization**

   i. Plot the loss function graphs for each parameter from steps 3

   ii. Display the reference point with a red dashed line and the loss value at learning completion with a green dashed line

6.3.2    Results and Discussion For Iris and Wine

Experimental results are shown in Figure 11 and 12.

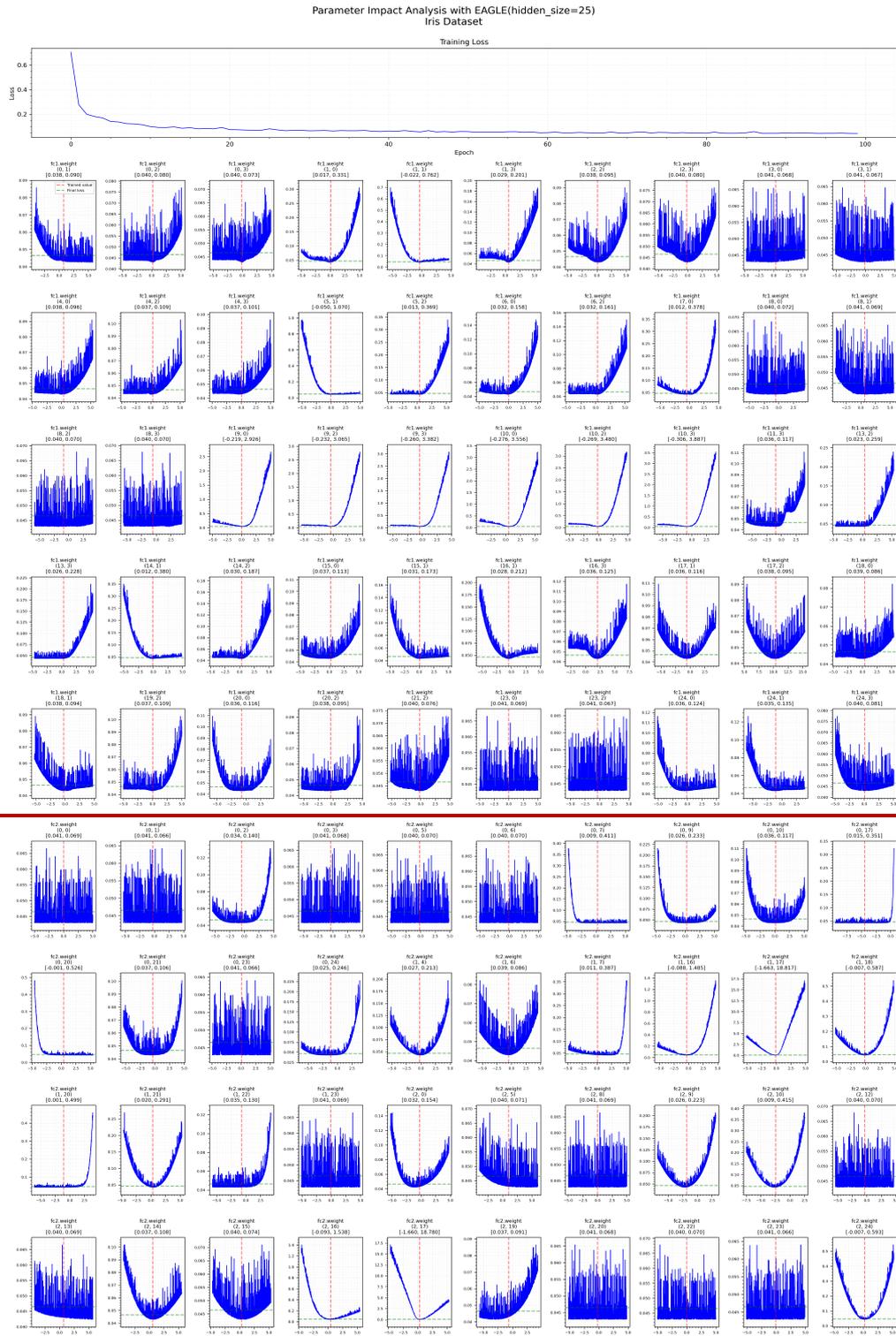

Figure 11: Local Shape of Loss Function (Iris)

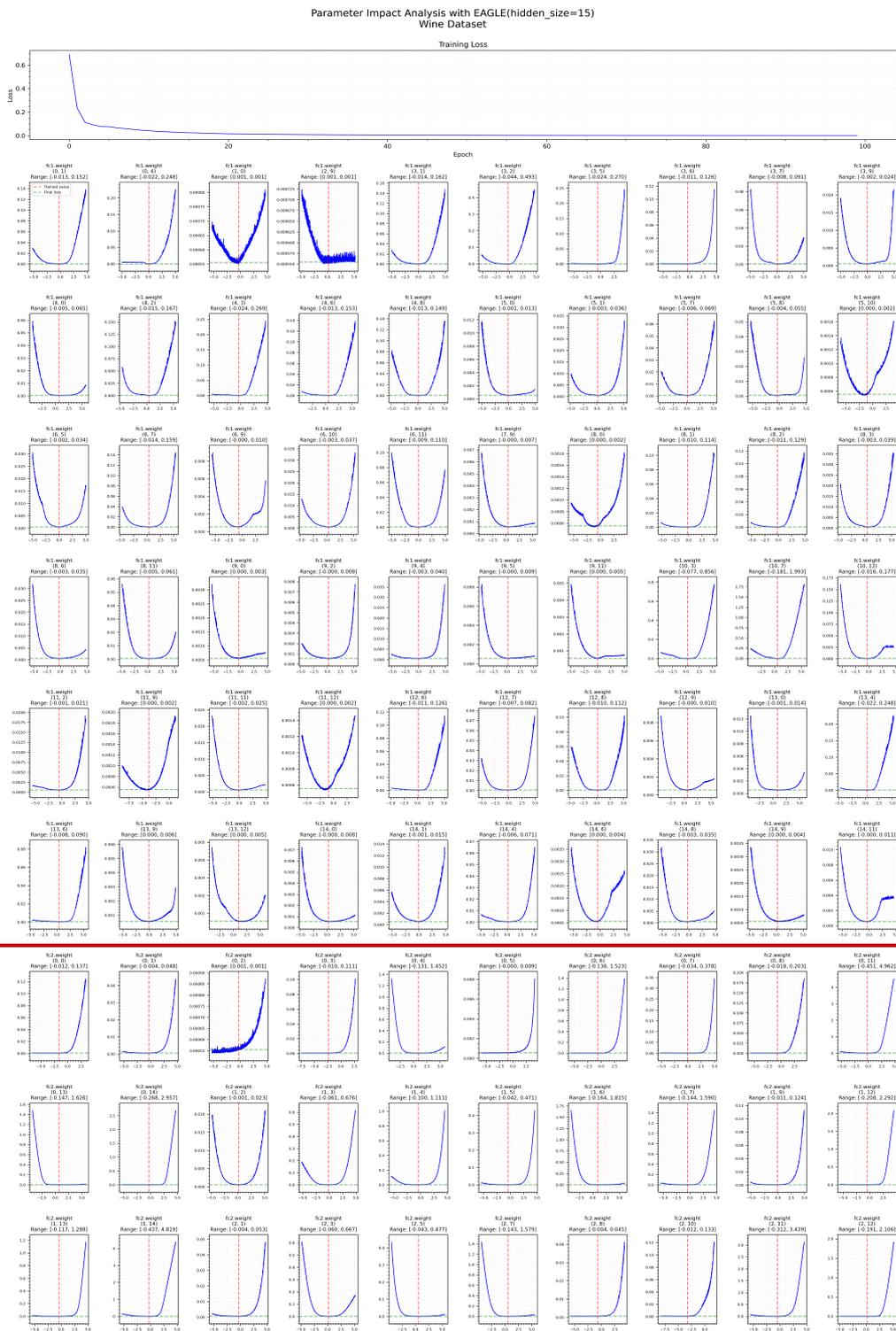

Figure 12: Local Shape of Loss Function (Wine)

We analyze the shape of loss functions obtained from experimental results from the following three perspectives.

- Shapes with Convexity

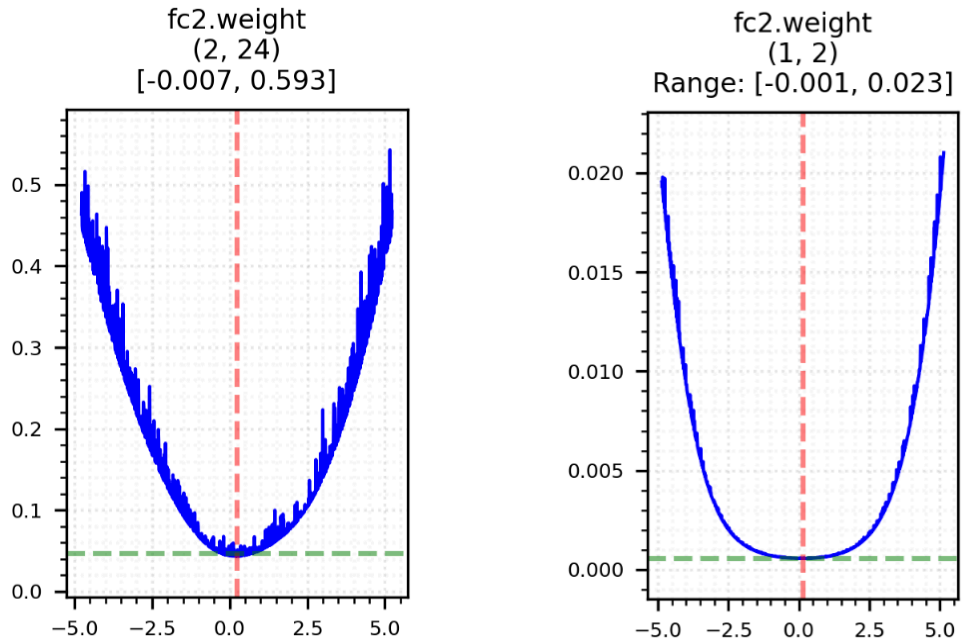

Figure 13: Shapes Approximating Convex (Left: Iris, Right: Wine)

As represented in Figure 13, multiple loss functions shown in Figures 11 and 12 were found to form curves that are convex around the reference point. This result supports the validity of our assumption that loss functions can be approximated as convex. This characteristic was particularly prominent in the Wine dataset. For parameters exhibiting such shapes, EAGLE update rule appears to function effectively, likely contributing to the rapid loss convergence in early training stages as shown in Section 6.1.

- Oscillatory Shapes

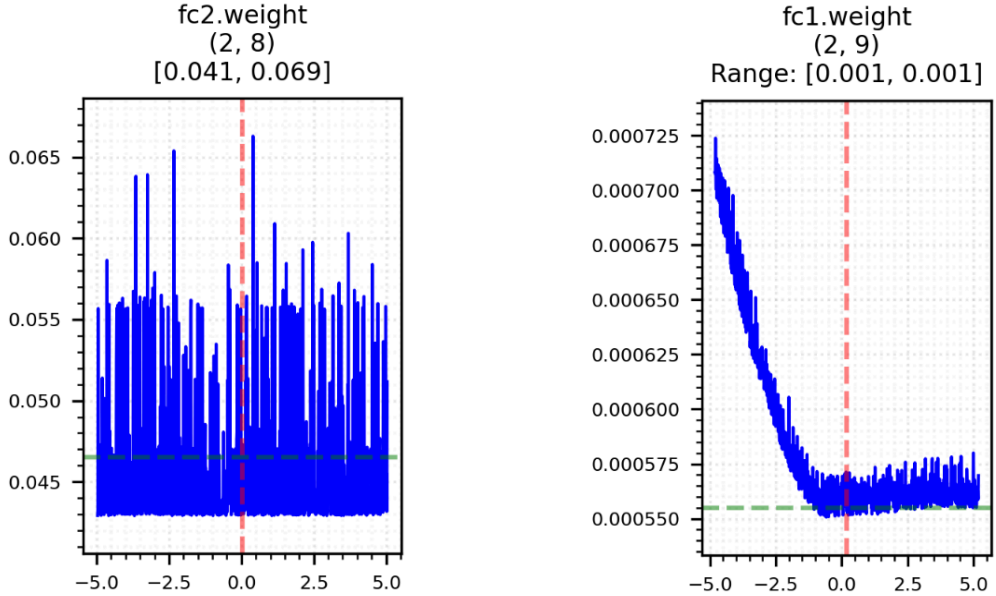

Figure 14: Shapes Exhibiting Oscillation (Left: Iris, Right: Wine)

As represented in Figure 14, multiple loss functions shown in Figures 11 and 12 were found to exhibit oscillatory behavior in response to parameter value changes, which appears to contradict our assumption that loss functions can be approximated as convex. However, since this experiment analyzes only the shape near the optimal solution, the overall shape of the loss function remains unconfirmed, and we cannot definitively conclude that it "contradicts the assumption."

Here, we consider the extent to which this oscillatory behavior affects model performance. Comparing Figures 13 and 14 with focus on the range of loss variations on the vertical axis, we observe a reduction from approximately 0.45 to 0.02 for the Iris dataset, and from approximately 0.02 to 0.000175 for the Wine dataset. Consequently, when adjusting the vertical scale of Figure 14 to match the convex shape graphs, the oscillating shapes can be interpreted as nearly flat. This indicates that the loss is close to its minimum value near the optimal solution, suggesting that parameter value fluctuations have minimal impact on the loss. In conclusion, parameters with such oscillatory loss function shapes likely have minimal impact on model performance through optimization, and precise parameter optimization may not be necessary.

However, even if the loss changes for individual parameters are small, when multiple parameters simultaneously influence the loss, their effects may accumulate and have non-negligible impact on the final loss value. Therefore, insufficient control of such oscillations near the optimal solution in EAGLE may have contributed to the decreased final convergence performance shown in Section 6.1.

Additionally, comparing Figures 11 and 12, such oscillatory loss functions were frequently observed in the Iris dataset but were less common in the Wine dataset. According to Tables 6 and 7, the final loss values across all optimizers were lower for the Wine dataset compared to the Iris dataset. This suggests that such oscillations may be a factor making optimization challenging for all optimizers, not just EAGLE.

- Shapes with Changing Signs of Second Derivatives

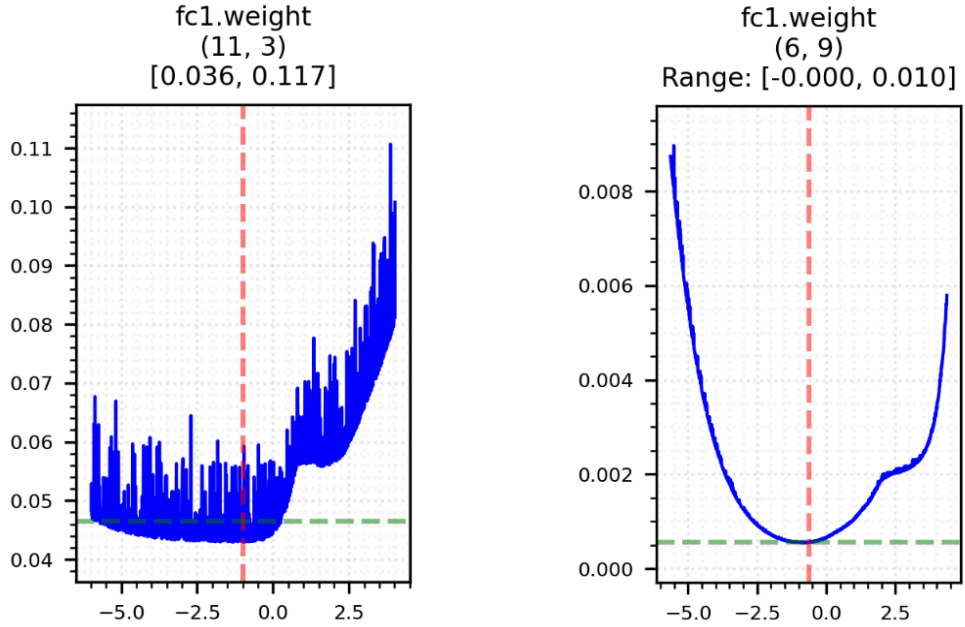

Figure 15: Shapes with Sign Changes in Second Derivatives (Left: Iris, Right: Wine)

As represented in Figure 15, among the loss functions shown in Figures 11 and 12, we confirmed the existence of loss functions with changing signs in second derivatives as discussed in Section 3.2, particularly in the Wine dataset. From the position of the final loss value (green dashed line) in Figure 15, we can observe that despite the presence of oscillations in Iris, parameters are properly updated to minimize loss even when second derivatives change signs. This demonstrates that the switching mechanism proposed in Section 3.2 functions effectively even with changes in second derivatives.

We summarize the above three analyses. This experiment confirmed the existence of loss functions that can be approximated as convex, and suggests that EAGLE functions effectively for these cases, achieving rapid convergence in early training stages. Additionally, we confirmed proper updates even with loss functions having second derivative sign changes, demonstrating the effectiveness of the proposed switching mechanism in our update rule. However, the handling of parameters with oscillatory loss functions remains insufficient, which was analyzed as a factor contributing to decreased final convergence performance.

### 6.3.3 RESULTS AND DISCUSSION FOR MNIST

Experimental results are shown in Figures 16, 17, 18, 19 and 20.

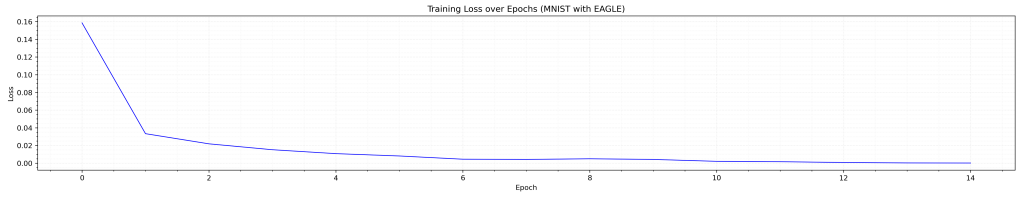

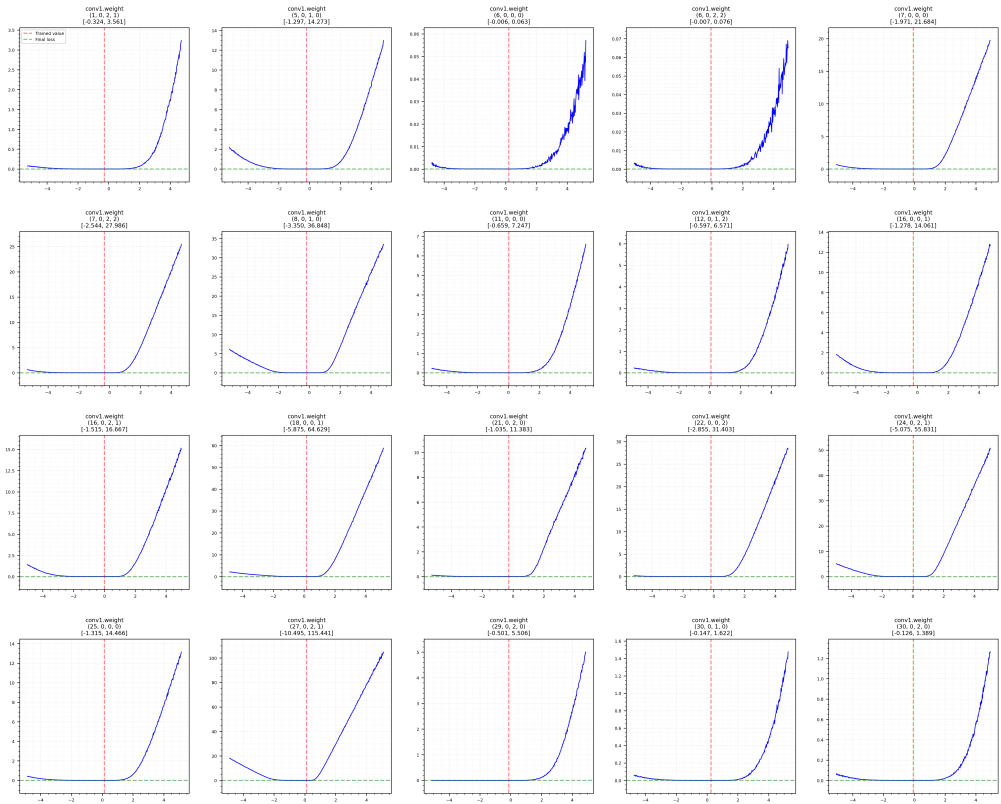

Figure 16: Changes in Training Loss During Reference Point Learning

Figure 17: Local Shape of Loss Function (MNIST: Convolutional Block 1)

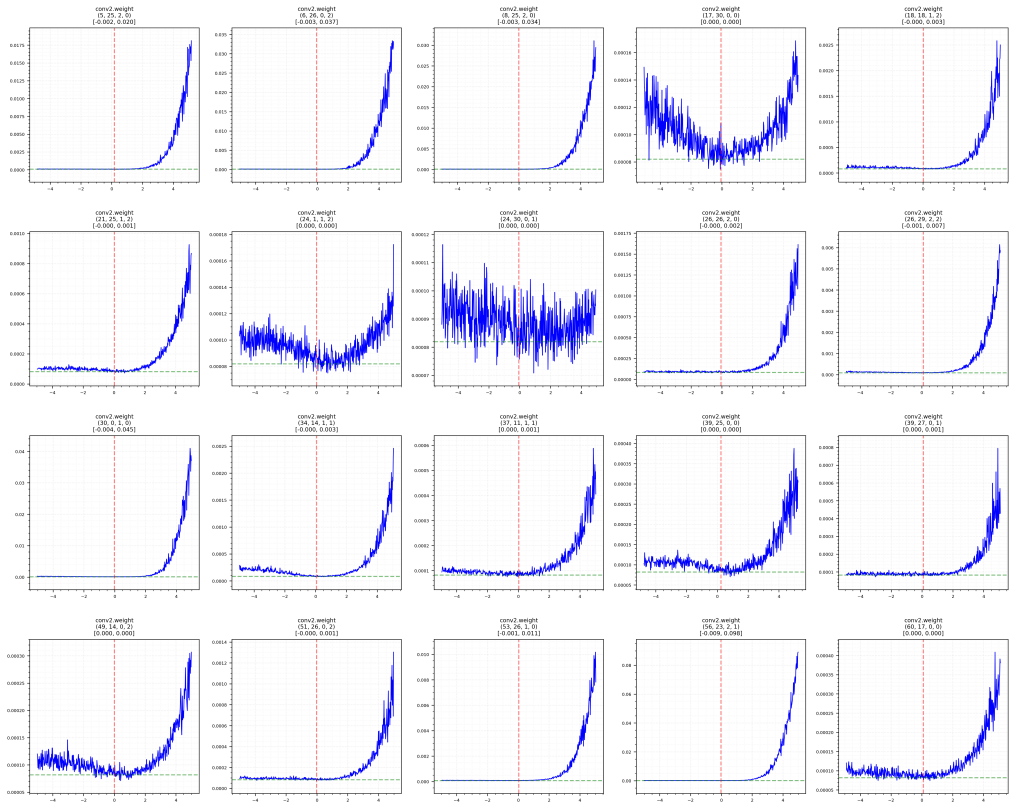

Figure 18: Local Shape of Loss Function (MNIST: Convolutional Block 2)

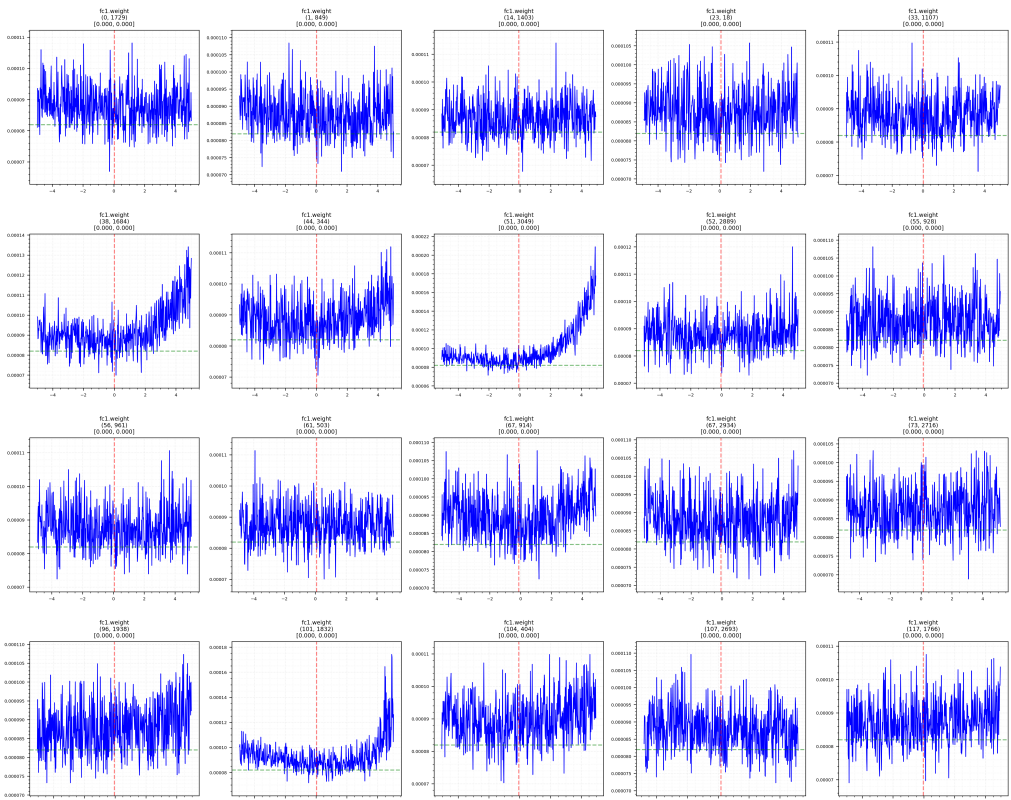

Figure 19: Local Shape of Loss Function (MNIST: First Fully Connected Layer)

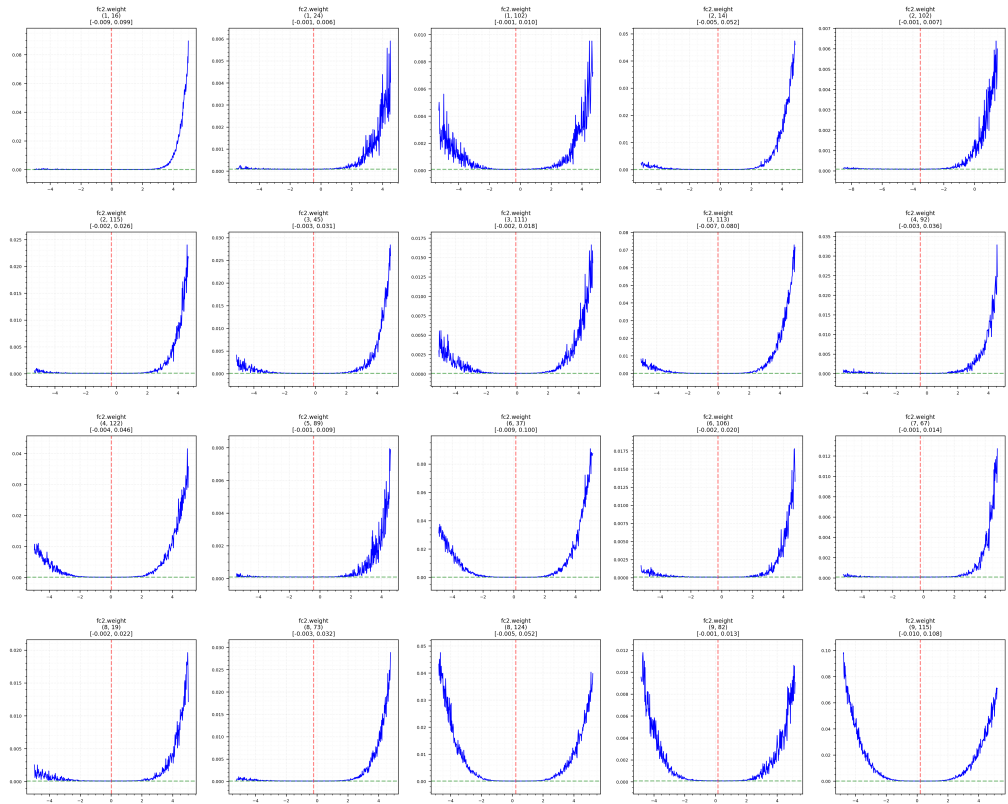

Figure 20: Local Shape of Loss Function (MNIST: Second Fully Connected Layer)

We analyze the lack of significant early-stage loss reduction in the MNIST dataset compared to that observed in the Iris and Wine datasets in Section 6.2, from the perspective of loss function shape characteristics.

- Oscillatory Shapes

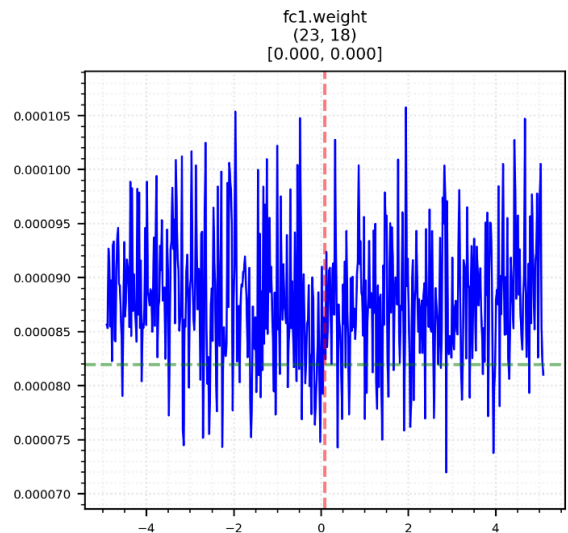

Figure 21: Examples of First Fully Connected Layer Parameters

We analyze the parameter updates in the first fully connected layer shown in Figure 19. In Figures 17, 18, 19, and 20, while we used the same sampling number for each layer in this experiment, the parameters in the first fully connected layer actually account for 95.2% of all parameters, and these parameters have oscillatory loss functions as shown in Figure 21. As discussed in Section 5.2, loss minimization by EAGLE is challenging for parameters with such oscillatory shapes. We conclude that this characteristic led to the absence of significant loss reduction in the early stages of training.

- Linear and Flat Shapes

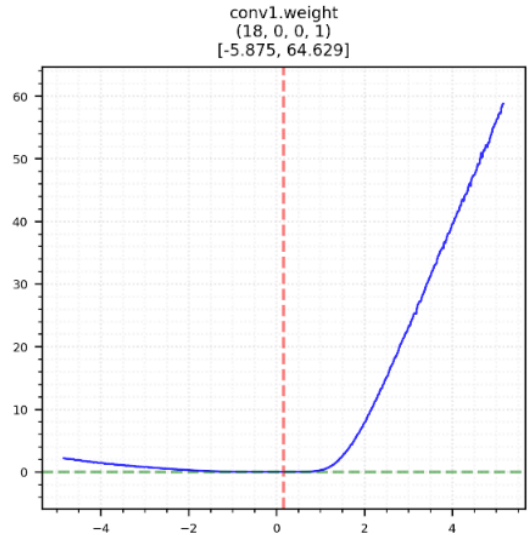

Figure 22: Examples of First Convolutional Block Parameters

We analyze the parameter updates in the first convolutional block shown in Figure 17. As shown in Figure 22, these parameter groups commonly exhibit linear (right side of optimal solution) or flat (left side of optimal solution) shapes near the optimal solution. As discussed in Section 3.1, these shapes result in extremely small gradient differences, which could lead to divergent update magnitudes when applying EAGLE update rule. However, Figure 17 shows no such divergence, and as indicated by the loss value at learning completion (green dashed line), updates successfully reduced the loss. This suggests that the switching mechanism functioned properly, and Adam update rule was primarily used for updating these parameters.

- Shapes with Convexity

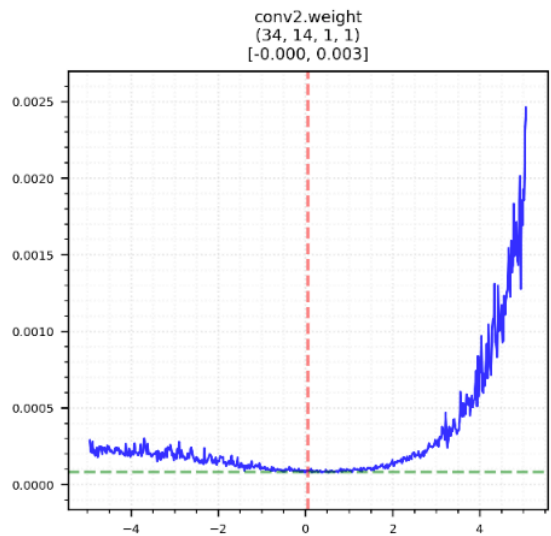

Figure 23: Examples of Second Convolutional Block Parameters

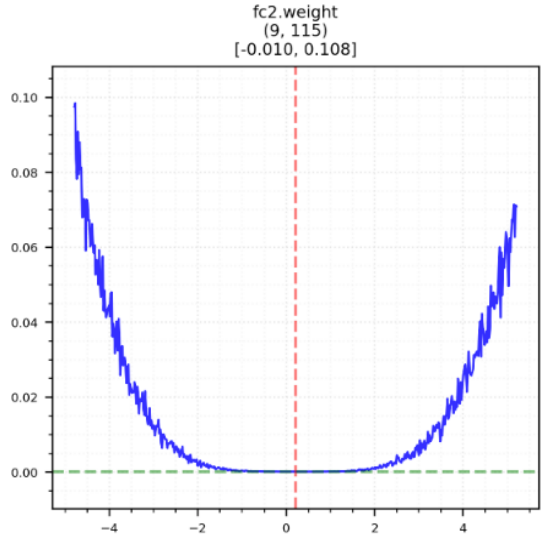

Figure 24: Examples of Second Fully Connected Layer Parameters

• Lower Threshold

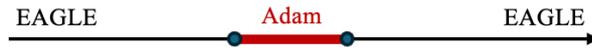

• Higher Threshold

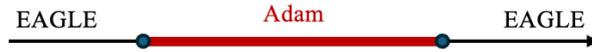

Figure 25: Changes in Update Rule Usage Ratio with Threshold Variation

As shown in Figures 18 and 20, the parameter groups in the second convolutional block and the second fully connected layer exhibit shapes with convexity, as illustrated in Figures 23 and 24. In these regions, EAGLE update rule appears to function effectively, contributing to accelerated learning.

## 6.4 EXPERIMENT: ANALYSIS OF EAGLE UPDATE RULE USAGE

### 6.4.1 EXPERIMENTAL SETTINGS

As explained in Chapter 3, the proposed EAGLE optimizer introduces a switching mechanism that adaptively switches between EAGLE update rule and Adam update rule. The gradient difference threshold explained in Section 3.1 is a crucial hyperparameter that directly affects the usage ratio as shown in Figure 25. In this experiment, we analyze the changes in usage rate of EAGLE update rule during the training process. Specifically, using the Iris, Wine, and MNIST datasets and their models from Sections 6.1 and 6.2, we evaluate the progression of training loss, changes in usage rates during early training and at completion, and average usage rate throughout training when varying this threshold across [1e-3, 7e-4, 4e-4, 1e-4].

Similar to Section 5.1, we conducted independent experiments using 5 randomly selected seed values (3 values for MNIST) from 1 to 10,000. We evaluated the experimental results using the following two metrics:

**Evaluation Metrics 1**: Time-series visualization of mean and standard deviation for Training Loss, Training Accuracy, Test Loss, and Test Accuracy across all seeds under varying thresholds. The standard deviation is represented by shaded regions around each curve.

**Evaluation Metrics 2**: Usage rate analysis of EAGLE update rule for each threshold value, presented in tabular form. The table shows three rates per threshold: Usage rate during early training (epoch 10) / Usage rate at learning completion (epoch 100) / Average usage rate across all epochs

### 6.4.2   RESULTS AND DISCUSSION FOR IRIS AND WINE

Experimental results are shown in Figures 26, 27, and Tables 9, 10.

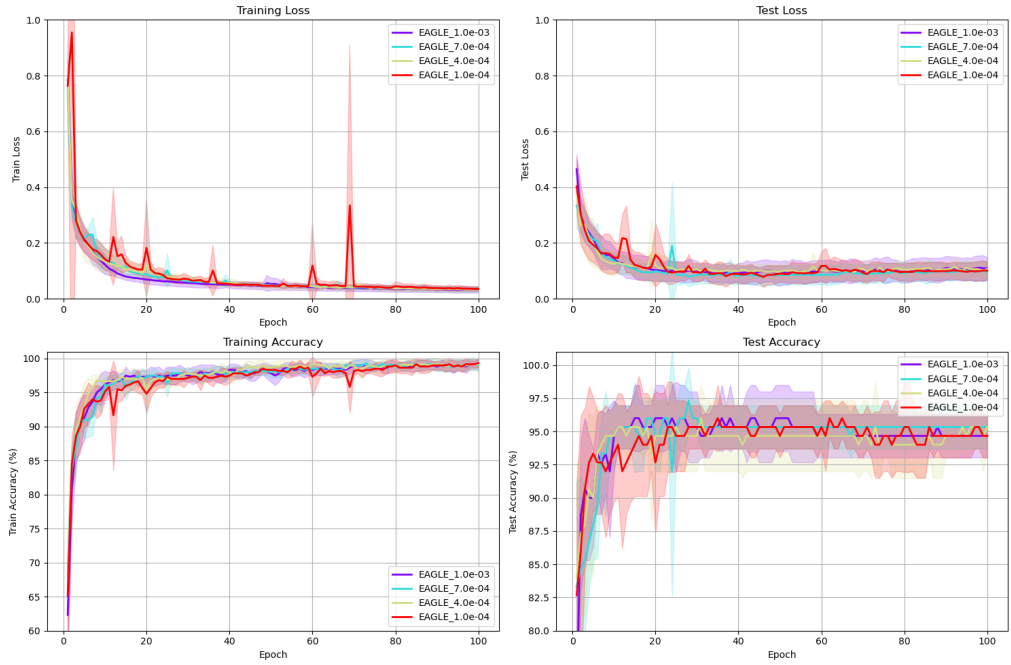

Figure 26: Evaluation Metrics 1 (Iris)

Table 9: Evaluation Metrics 2 (Iris)

| threshold | | 3958 | 1069 | 1917 | 5609 | 8860 | average |
|---|---|---|---|---|---|---|---|
| **0.001** | **10** | 49.88 | 46.55 | 44.40 | 51.23 | 41.93 | **46.80** |
| | **100** | 36.95 | 26.66 | 39.41 | 43.47 | 34.05 | **36.11** |
| | **ave** | 45.49 | 39.65 | 41.83 | 49.21 | 37.67 | **42.77** |
| **0.0007** | **10** | 47.60 | 51.11 | 44.52 | 50.25 | 51.05 | **48.91** |
| | **100** | 43.41 | 39.53 | 48.77 | 35.53 | 28.33 | **39.11** |
| | **ave** | 43.72 | 46.71 | 45.00 | 46.74 | 49.95 | **46.42** |
| **0.0004** | **10** | 66.69 | 63.36 | 54.86 | 60.90 | 58.13 | **60.79** |
| | **100** | 56.03 | 54.13 | 47.23 | 37.62 | 50.00 | **49.00** |
| | **ave** | 54.55 | 56.93 | 56.35 | 55.36 | 54.53 | **55.54** |
| **0.0001** | **10** | 63.92 | 57.33 | 58.87 | 63.05 | 46.18 | **57.87** |
| | **100** | 52.34 | 45.38 | 50.68 | 41.44 | 42.43 | **46.45** |
| | **ave** | 53.56 | 52.26 | 57.43 | 52.56 | 42.09 | **51.18** |

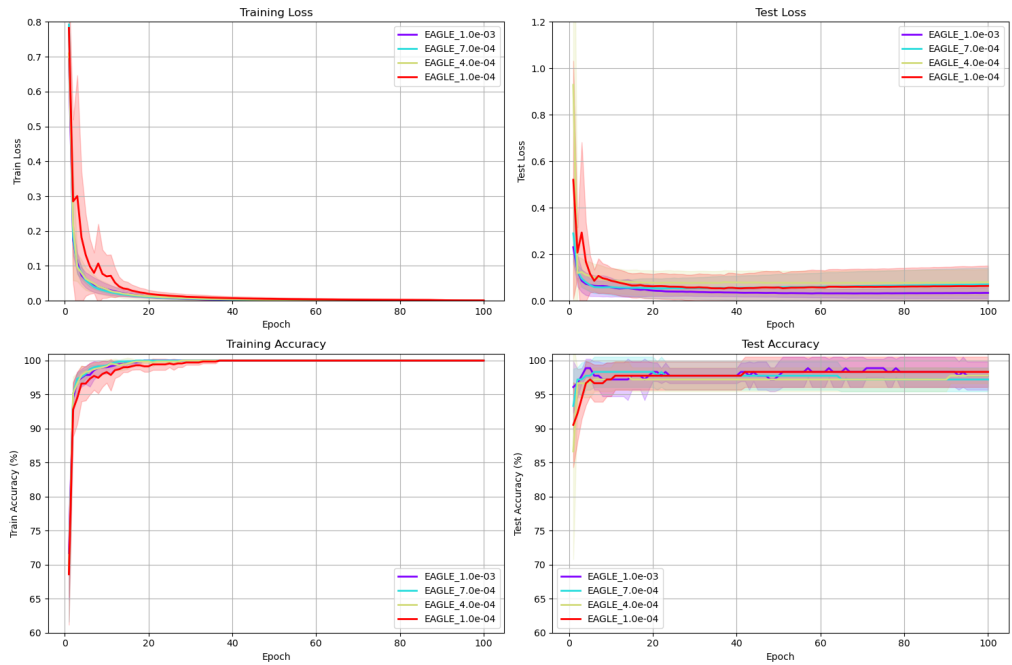

Figure 27: Evaluation Metrics 1 (Wine)

Table 10: Evaluation Metrics 2 (Wine)

| threshold | | 5562 | 9101 | 8474 | 3174 | 4748 | average |
|---|---|---|---|---|---|---|---|
| **0.001** | **10** | 49.83 | 47.93 | 51.38 | 54.31 | 55.86 | **51.86** |
| | **100** | 1.03 | 1.34 | 2.20 | 6.16 | 0.22 | **2.19** |
| | **ave** | 19.40 | 20.67 | 29.78 | 30.24 | 21.12 | **24.24** |
| **0.0007** | **10** | 58.74 | 60.03 | 55.47 | 62.36 | 58.23 | **58.97** |
| | **100** | 6.20 | 2.63 | 9.52 | 9.60 | 8.91 | **7.37** |
| | **ave** | 29.57 | 28.22 | 34.43 | 36.89 | 30.06 | **31.83** |
| **0.0004** | **10** | 60.90 | 59.95 | 55.21 | 66.24 | 62.58 | **60.98** |
| | **100** | 14.43 | 6.59 | 9.09 | 7.92 | 32.21 | **14.05** |
| | **ave** | 36.97 | 34.25 | 39.28 | 36.57 | 47.71 | **38.96** |
| **0.0001** | **10** | 66.49 | 55.00 | 57.02 | 67.36 | 64.86 | **62.15** |
| | **100** | 42.94 | 23.00 | 40.74 | 43.11 | 45.26 | **39.01** |
| | **ave** | 54.19 | 47.16 | 50.12 | 57.57 | 56.66 | **53.14** |

We analyze the graphs and tables obtained from the experimental results from the following two perspectives.

- Changes in EAGLE Update Rule Usage Rate

  ➢ Increase in EAGLE Update Rule Usage with Decreasing Threshold

    As shown in Table 10, for the Wine dataset, across all seeds on average, a consistent trend was observed where lower thresholds led to higher usage rates of EAGLE update rule at epoch 10, epoch 100, and in overall average. This is a reasonable result, as shown in Figure 25, since lower thresholds increase the regions where EAGLE update rule is applied.

    However, in the Iris dataset, as shown in Table 9, the usage rate decreased when the threshold changed from 0.0004 to 0.0001, contradicting the results observed in Table 10. Detailed examination of individual seeds revealed different tendencies in this threshold range, with three seeds (1069, 3958, 8860) showing decreased usage and two seeds (1917, 5609) showing increased usage. This indicates unstable update rule selection at lower thresholds, likely due to the abundance of parameters with oscillatory loss function shapes in the Iris dataset, as shown in Figure 14 of Section 6.2.

  ➢ Decrease in Usage Rate from Epoch 10 to Epoch 100

    Regarding changes across learning stages, Tables 9 and 10 show a consistent decrease in usage rates from epoch 10 to epoch 100 for both datasets. This result indicates that EAGLE update rule is used relatively more frequently in early training stages, demonstrating its effectiveness in achieving rapid loss convergence during early learning, which is the objective of this research.

    The decrease in usage rate was particularly pronounced in the Wine dataset. While the Iris dataset showed an average decrease of approximately 11%, the Wine dataset exhibited approximately 43% decrease, showing about four times greater reduction rate. Notably, with a threshold of 0.001 in the Wine dataset, the usage rate decreased by approximately 96%, from 51.86% to 2.19%. As discussed in Section 6.2, this indicates that due to the Wine dataset's relatively well-structured convex loss shapes, EAGLE update rule was effectively used in early training, followed by appropriate transition to Adam update rule in later stages.

- Learning Stability

  Within the threshold range set in this experiment, as shown in Figures 26 and 27, no significant impact on learning outcomes was observed except for the 0.0001 case, demonstrating stable learning. However, at the minimum value of 0.0001, oscillations were observed in the learning process, particularly in the Iris dataset, indicating potential convergence instability. This suggests signs of update divergence due to excessive selection of EAGLE update rule, indicating possible inappropriate updates at even lower values.

These results confirm that EAGLE possesses certain robustness to the threshold hyperparameter, and for the datasets and models in this experiment, moderate threshold values (0.0004-0.0007) achieve a good balance between stable learning and appropriate usage of EAGLE update rule.

### 6.4.3 RESULTS AND DISCUSSION FOR MNIST

Experimental results are shown in Figure 28 and Table 11.

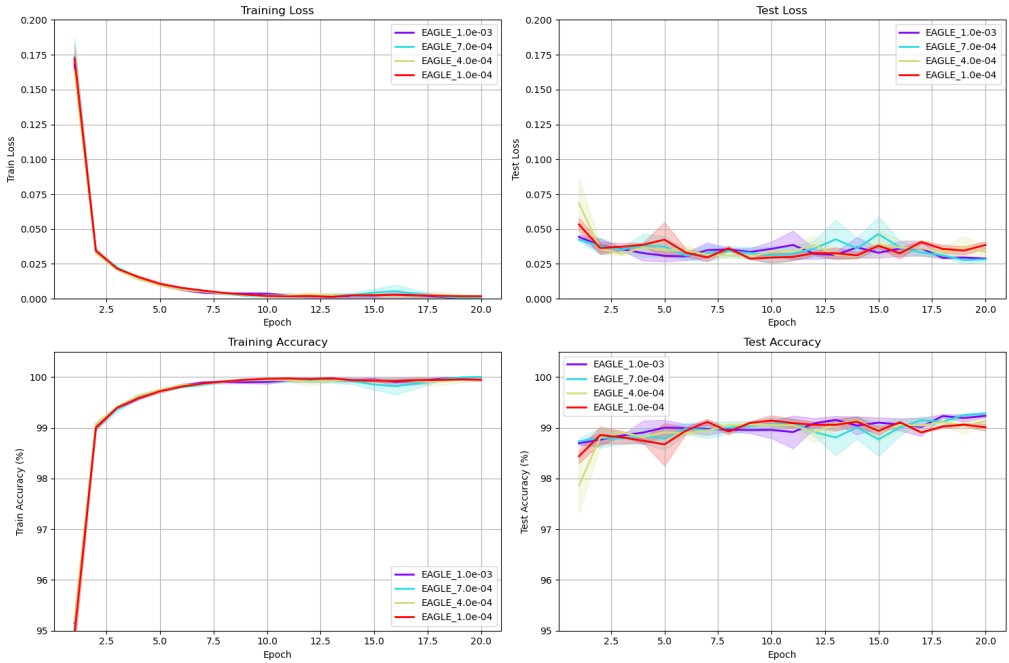

Figure 28: Evaluation Metrics 1 (MNIST)

Table 11: Evaluation Metrics 2 (MNIST)

| threshold | | 7381 | 4014 | 5199 | average |
|---|---|---|---|---|---|
| **0.001** | **2** | 0.42 | 0.40 | 0.50 | **0.4400** |
| | **20** | 0.00 | 0.00 | 0.01 | **0.0033** |
| | ave | 0.19 | 0.21 | 0.25 | **0.2167** |
| **0.0007** | **2** | 1.15 | 0.93 | 0.96 | **1.1033** |
| | **20** | 0.01 | 0.00 | 0.01 | **0.0067** |
| | ave | 0.49 | 0.39 | 0.48 | **0.4533** |
| **0.0004** | **2** | 2.80 | 2.97 | 2.68 | **2.8167** |
| | **20** | 0.00 | 0.28 | 0.31 | **0.1967** |
| | ave | 1.07 | 1.15 | 1.19 | **1.1367** |
| **0.0001** | **2** | 21.02 | 21.38 | 21.81 | **21.4033** |
| | **20** | 4.72 | 3.19 | 2.16 | **3.357** |
| | ave | 8.41 | 8.53 | 8.57 | **8.5033** |

- Behavioral Similarity with Adam Optimizer

  From Figure 9, which shows the experimental results of Section 6.1, we can observe that the behavior throughout the learning process is similar to that of Adam optimizer. The results of this experiment explain the reason for this similarity. The gradient difference threshold set in Section 6.1's experiment was 0.0005, and focusing on the results for 0.0004, which is closest to 0.0005 among the thresholds used in this experiment, Table 11 shows that the average usage rate of EAGLE update rule was approximately 1.14%. This result indicates that the majority of parameter updates were executed by Adam update rule, explaining the behavioral similarity with Adam. However, the 1.14% of updates by EAGLE update rule achieved the early-stage acceleration shown in Section 6.1's experiments. This result demonstrates that EAGLE update rule can achieve effective parameter updates even when applied locally.

# 7 CONCLUSION AND FUTURE WORK

In this research, we proposed EAGLE, a parameter optimization method utilizing local curvature of loss functions obtained from gradient variations between consecutive steps. By combining this with a switching mechanism based on gradient difference thresholds and loss function shapes as a practical extension, we aimed to achieve both rapid loss convergence in early training stages and improved overall training stability. From four experiments, we obtained the following conclusions:

- In fully connected neural network models using Iris and Wine datasets, and a CNN model using the MNIST dataset, EAGLE accelerated loss convergence in early training stages compared to conventional methods while maintaining stable learning thereafter. Particularly with the Wine dataset, approximately 75% reduction in loss value compared to Adam was achieved in early training. However, as learning progressed, the loss difference diminished, revealing an issue where the final loss value was higher than conventional methods.

- To verify our assumed convexity, we analyzed the loss function shapes of Iris, Wine, and MNIST datasets and their models. This confirmed the existence of loss functions that can be approximated as convex, and we concluded that EAGLE update rule's effective function on these shapes enabled rapid loss convergence in early training stages. However, we also identified loss functions with oscillatory shapes, where these parameters were not fully minimized. While individual parameter effects were small, their cumulative impact led to decreased final convergence performance.

- To analyze changes in EAGLE update rule usage rates through the switching mechanism, we varied the gradient difference threshold. This confirmed adaptive update rule selection according to threshold changes and dataset/model characteristics. Particularly in the Wine dataset, we observed desired behavior where EAGLE update rule was utilized in early training before transitioning to Adam update rule as learning progressed. In the MNIST dataset, despite relatively low usage rates, early-stage acceleration was achieved, demonstrating that EAGLE can perform effective updates even with localized usage. Additionally, certain robustness to threshold settings was confirmed, particularly around 0.0005 used throughout the experiments, showing balanced achievement of EAGLE update rule's rapid loss convergence and Adam update rule's stable learning.

Future challenges regarding these conclusions include:

- The issue of decreased final convergence performance due to cumulative effects of individual parameters exists. To properly control this impact, introducing momentum terms to suppress oscillations or complete switching to Adam in later learning stages could be considered.

- While currently setting 0.0005 as the default threshold value, appropriate value selection based on dataset and model characteristics is necessary. Specifically, mechanisms for dynamically adjusting thresholds as learning progresses or methods for automatically setting thresholds based on model scale could be considered.

- The current implementation requires calculating gradient variations individually for

each parameter, increasing memory usage. Implementation methods to improve memory efficiency need to be examined.

- While the current implementation introduces branching based on gradient difference magnitude, alternatively, introducing a switching mechanism that prevents update magnitude divergence by setting thresholds for the update magnitude itself needs to be considered.

APPENDIX

We conduct the same experiments as in Section 6.1 using the CIFAR-10 dataset to evaluate the performance of EAGLE. The CIFAR-10 dataset is used for object recognition tasks. Each object is represented as a 32×32 pixel RGB image and classified into 10 classes: airplane, automobile, bird, cat, deer, dog, frog, horse, ship, and truck. We adopt a pre-trained Vision Transformer (ViT-H/14) as our model. ViT-H/14 is a large Transformer model pre-trained on the ImageNet dataset, with a patch size of 14×14 pixels and an input image size of 224×224 pixels. In this research, we resize CIFAR-10 images to 224×224 and use the model by fine-tuning the final layer for CIFAR-10's 10-class classification. For the experimental method, due to the large scale of the ViT-H/14 model, we conducted a single experiment with one seed.

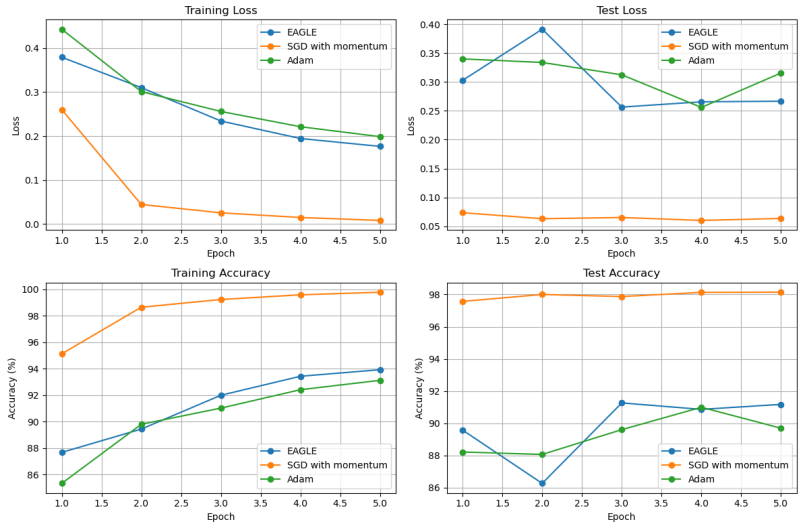

Figure : Changes in Training Loss/Training Accuracy

Table : Training Loss/Training Accuracy per Epoch

|  | 1 | 2 | 3 | 4 | 5 |
|---|---|---|---|---|---|
| EAGLE | 0.3793 | 0.3102 | 0.2343 | 0.1945 | 0.1766 |
| Adam | 0.4430 | 0.3012 | 0.2560 | 0.2213 | 0.1988 |
| SGD | **0.2603** | **0.0445** | **0.0256** | **0.0150** | **0.0083** |

Table : Usage Rate of EAGLE Update Rule per Epoch

| Num_epoch | 1 | 2 | 3 | 4 | 5 |
|---|---|---|---|---|---|
| Usage Rate | 0.1257 | 0.1249 | 0.0751 | 0.0664 | 0.0576 |

From  Figure, training loss decreases monotonically throughout learning, confirming that EAGLE maintains convergence capability even with Transformer models. Additionally, similar to the MNIST experiment results, it demonstrated convergence performance comparable to Adam. In comparison with SGD with momentum, while the training loss difference was minimal at epoch 1, it reached its maximum at epoch 2. This difference did not significantly decrease with increasing epochs, and the characteristic rapid loss convergence in early training stages observed in previous experiments was not sufficiently apparent. We analyze these experimental results from the following two perspectives.

- Comparison with Adam

  The usage rate of EAGLE update rule in this experiment averaged approximately 0.089%, which is even lower compared to the MNIST experiment (approximately 1.14%). The fact that most updates were executed by Adam update rule likely explains the similarity in convergence performance with Adam. However, at epoch 1, EAGLE's training loss was 14.38% lower than Adam's, suggesting that EAGLE update rule functions effectively in local regions.

- Comparison with SGD and Lack of Early Convergence

  While we should analyze the loss function shape as in Section 6.3, in this experiment, a single training session of 5 epochs with three optimization methods involved 2,466,340,455,100 parameter updates, making shape analysis impractical. However, Transformer models are known to have complex nonlinear structures, suggesting relatively fewer regions with convex loss shapes. This characteristic likely prevented EAGLE update rule from fully demonstrating its effectiveness, possibly explaining why the early convergence observed in conventional FCNNs and CNNs did not manifest. Nevertheless, the adaptive learning rate characteristic of Adam update rule appears well-suited to this model, enabling reasonable convergence performance.

These experimental results indicate that optimizer performance depends on model architecture, demonstrating that optimizers with adaptive learning rates, such as EAGLE and Adam, may not always be the optimal choice. This suggests the necessity of selecting optimizers with careful consideration of model architectural characteristics.